\documentclass[10pt,twocolumn,letterpaper]{article}

\usepackage[pagenumbers]{cvpr} 

\usepackage{graphicx}
\usepackage{amsmath}
\usepackage{amssymb}
\usepackage{booktabs}
\usepackage{enumitem}
\usepackage{cuted}
\usepackage{multirow}
\usepackage{xcolor}
\usepackage[table]{xcolor}
\usepackage[ruled,vlined]{algorithm2e}
\usepackage{tocloft}
\usepackage{titletoc}

\SetKwInput{KwInput}{Input}
\SetKwInput{KwOutput}{Output}
\SetKwComment{Comment}{$\triangleright$}{}

\definecolor{cvprblue}{rgb}{0.21,0.49,0.74}
\usepackage[pagebackref,breaklinks,colorlinks,allcolors=cvprblue]{hyperref}

\newcommand{\paravspace}{\vspace{-11pt}}

\def\confName{CVPR}
\def\confYear{2026}

\title{\textcolor{black}{Native and Compact Structured Latents for 3D Generation}\thanks{Open-source project; see our \href{https://microsoft.github.io/TRELLIS.2}{project page} for code, model, and data.}}

\makeatletter
\renewcommand*{\@fnsymbol}[1]{\ensuremath{\ifcase#1\or *\or \star\or \dagger\or \ddagger\or
		\mathsection\or \mathparagraph\or \|\or **\or \dagger\dagger
		\or \ddagger\ddagger \else\@ctrerr\fi}}
\makeatother

\author{
    Jianfeng Xiang$^{1,2}$\thanks{Work done during internship at Microsoft Research} \quad  Xiaoxue Chen$^{1\star}$ \quad  Sicheng Xu$^{2}$ \quad Ruicheng Wang$^{3,2\star}$ \quad Zelong Lv$^{3,2\star}$ \\
    Yu Deng$^{2}$ \quad Hongyuan Zhu$^{4}$ \quad Yue Dong$^{2}$\quad Hao Zhao$^{1}$ \quad Nicholas Jing Yuan$^{4}$ \quad Jiaolong Yang$^{2}$\thanks{Corresponding author} \\
	$^1${Tsinghua University} \quad $^2${Microsoft Research} \quad $^3${USTC} \quad $^4${Microsoft AI}\\
    \url{https://microsoft.github.io/TRELLIS.2}
}

\begin{document}
\maketitle

\begin{strip}
	\vspace{-36pt}
	\centering
	\includegraphics[width=1\textwidth]{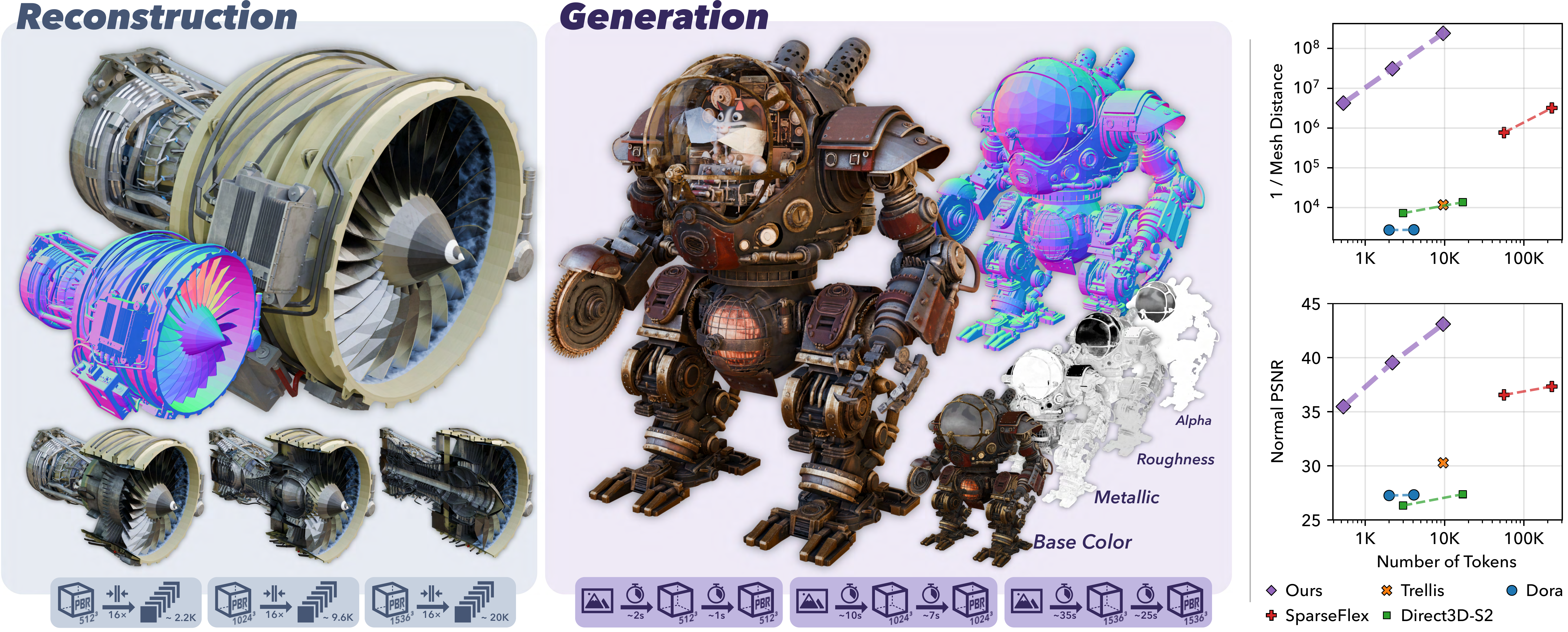}
    \vspace{-8pt}
    \captionsetup{type=figure,font=small,position=top}
    \caption{
    \textbf{Left:} A 1536³ asset reconstruction by our method. Despite the high compactness of the latents (see token counts below), it faithfully recovers extremely fine geometric and material details, supports arbitrary topology, and preserves enclosed structures (shown in the second row). \textbf{Middle:}  A 1536³ 3D asset generated in about one minute ($\sim$35s for shape and $\sim$25s for texture; see more runtime in bottom row). Building on our latents, the generator efficiently produces high-quality PBR-textured assets, delivering intricate geometric detail and realistic materials across open-domain inputs. \textbf{Right:} Latent representation comparison on shape reconstruction. Our method achieves much higher fidelity with modest token number, pushing the frontier of both quality and compactness. \textbf{\emph{(Best viewed with zoom.)}}
    }
    \label{fig:teaser}
    \vspace{-10pt}
\end{strip}

\begin{abstract}
Recent advancements in 3D generative modeling have significantly improved the generation realism, yet the field is still hampered by existing representations, which struggle to capture assets with complex topologies and detailed appearance.
This paper presents an approach for learning a structured latent representation from native 3D data to address this challenge. At its core is a new sparse voxel structure called O-Voxel, an omni-voxel representation that encodes both geometry and appearance. O-Voxel can robustly model arbitrary topology, including open, non-manifold, and fully-enclosed surfaces, while capturing comprehensive surface attributes beyond texture color, such as physically-based rendering parameters. Based on O-Voxel, we design a Sparse Compression VAE which provides a high spatial compression rate and a compact latent space. We train large-scale flow-matching models comprising 4B parameters for 3D generation using diverse public 3D asset datasets. Despite their scale, inference remains highly efficient. Meanwhile, the geometry and material quality of our generated assets far exceed those of existing models. We believe our approach offers a significant advancement in 3D generative modeling.
\end{abstract}

\section{Introduction}

3D generative modeling has progressed at an unprecedented rate recently, spurred by innovations in latent 3D representation design and the integration of large latent learning and generative models~\cite{zhang2024clay,xiang2025structured,chen2025dora,he2025sparseflex,wu2025direct3d}. These advancements have dramatically enhanced both reconstruction fidelity and generation realism, bringing 3D content creation closer to real-world deployment and industrial applications.

Despite the progress, the field still lacks fundamental representations that can both faithfully capture the \emph{full-spectrum information} of \emph{arbitrary} 3D assets and effectively be processed into latents with neural networks. Recent large 3D generation models~\cite{zhang2024clay,xiang2025structured,chen20243dtopia,chen2025dora,wu2025direct3d,he2025sparseflex,li2025sparc3d} predominantly leverage \emph{iso-surface fields} (\eg, signed distance function, Flexicubes~\cite{shen2023flexicubes}) to represent geometry, which have \emph{intrinsic limitations in handling open surfaces, non-manifold geometry, and enclosed interior structures}. Moreover, most existing works focus on 3D shape generation while \emph{neglecting the appearance and material information} inherent in 3D assets that are fundamentally correlated with shape. \cite{xiang2025structured} introduces a structured 3D latent (SLAT) representation that jointly models geometry and appearance, but its reliance on multiview 2D image feature input and pure rendering-based supervision leads to deficiencies in capturing complex structures and materials.

In this work, we introduce a new \emph{``field-free"} sparse voxel structure termed \emph{O-Voxel}. O-Voxel is an omni-voxel representation that encodes both geometry and appearance, serving as a nexus between mesh assets and neural networks. The ``omnipotence" of this structure not only lies in its integrated approach to modeling geometry and appearance but also in its robust  capacity to handle their inherent complexity. For \emph{geometry}, it can handle arbitrary topology including open, non-manifold, and fully-enclosed surfaces, unlike existing field-based methods.
By introducing a flexible dual grid structure corresponding to the primal sparse voxels, 
it can accommodate complex topology without lossy data preprocessing and preserve sharp edge features and normal discontinuities for accurate geometry reproduction. For \emph{appearance}, it can capture arbitrary surface attributes beyond mere texture color. In this work, we implement physically-based rendering (PBR) parameters aligned with surface geometry to enable re-lighting capability. In particular, it incorporates material opacity, allowing it to handle translucent surfaces—a capability not present in previous methods.

Another notable advantage of O-Voxel is its instant bidirectional conversions to and from raw 3D assets. The processes are both optimization-free and rendering-free. Transforming a mesh into this structure takes only a few seconds on a single CPU, while reconstructing surfaces and materials from it completes within tens of milliseconds.

Given the O-Voxel representation, we design a sparse 3D variational autoencoder to learn a proper latent space. This leads to a \emph{native} structured latent space compared to that in \cite{xiang2025structured} which is built from multiview 2D information. Furthermore, we aim for a \emph{compact} latent space to support efficient and high-resolution 3D generation. 
Utilizing a residual autoencoding design~\cite{chen2024deep} applied to  sparse voxel structure, our VAE achieves a \emph{16$\times$ spatial downsampling}, a high ratio not seen in prior voxel-based methods. Our approach encodes a fully textured asset with \emph{$\textrm{1024}^\textrm{3}$ resolution into only $\sim$9.6K latent tokens} with negligible perceptual degradation upon reconstruction. Experiments show that our reconstruction quality surpasses prior methods by a wide margin, albeit using substantially fewer tokens. In addition to enabling  high-resolution 3D generation, our compact latent space also facilitates the scaling of the generative model.

We train large flow-matching generative models in the learned latent space for image-to-3D generation. 
The models contain about \emph{4 billion} parameters in total and are trained on diverse public 3D asset datasets. Despite their scale, \emph{inference remains highly efficient}: it takes only $\sim$3s for $512^3$ fully-textured assets generation, $\sim$17s for the $1024^3$ resolution, and $\sim$60s for the $1536^3$ resolution on a NVIDIA H100 GPU, which are significantly faster than existing large 3D generation models. Meanwhile, the geometry and material quality of our generated assets far exceed those of existing models, as demonstrated in our experiments. 
\textbf{\emph{All our model, code, and dataset will be publicly released to facilitate reproduction and further research.}}

\section{Related Work}

\paragraph{3D Representations for Generation.}
Effectively representing geometry and appearance for neural network processing is a key challenge in 3D generation.
Early works adopted implicit fields or their discretized structures to represent shape, such as occupancy fields~\cite{mescheder2019occupancy} and Signed Distance Functions (SDF)~\cite{park2019deepsdf,deng2021deformed,hui2022neural,zheng2023lasdiffusion}.
NeRF~\cite{mildenhall2021nerf,tang2023volumediffusion,muller2023diffrf}  integrates geometry and appearance in a radiance field, yielding realistic rendering but suffering from low geometry quality and heavy sampling costs.
Unstructured 3D representations, including meshes~\cite{nash2020polygen,chen2024meshanything}, point clouds~\cite{nichol2022point,luo2021diffusion,zhou20213d}, and  Gaussians~\cite{kerbl20233d,yu2024mip,zhang2024gaussiancube,he2024gvgen}, offer explicit 3D representations but lack structural regularity, posing challenges for network processing and latent compression.
Recent works have introduced structured representations tailored for 3D generation~\cite{he2025sparseflex,li2025sparc3d}. They combine field-based iso-surface modeling with sparse voxels to achieve high-resolution geometry. However, their reliance on field-based primitives limits their capability to represent open or non-manifold surfaces and they do not handle appearance, unlike our approach.

\paravspace
\paragraph{Latent 3D Representations.}
Recent advances in 3D generation have increasingly shifted from using explicit geometric representations to compact latent spaces, analogous to those used in the 2D domain~\cite{rombach2022high}.
Various latent formulations have been explored, including latent point clouds~\cite{lan2024ga,yang2024atlas,chen20243dtopia,vahdat2022lion}, volumetric or hierarchical grids~\cite{ntavelis2023autodecoding,xiong2025octfusion,ren2024xcube,meng2025lt3sd}, and triplane embeddings~\cite{wu2024direct3d,lan2024ln3diff,gupta20233dgen}.
Among these, two latent paradigms have emerged as the dominant choices for recent large 3D generation models.
The first is the \emph{unstructured latent}, inspired by the Perceiver-style architectures~\cite{jaegle2021perceiver}, where 3D data are encoded as unordered  feature vectors.
Representative works include \cite{zhang20233dshape2vecset,zhao2024michelangelo} and the follow-up methods~\cite{jun2023shap,zhang2024clay,li2024craftsman,li2025step1x,hunyuan3d2025hunyuan3d}.
These methods can achieve strong compression but are typically constrained by reconstruction fidelity.
The second category is the \emph{structured latent} built upon sparsity priors, exemplified by \cite{xiang2025structured} and its extensions~\cite{li2025sparc3d,he2025sparseflex}.
Such representations yield high geometric accuracy but require a larger number of latent tokens, which reduces compression efficiency.
Some efforts~\cite{wu2025direct3d,chen2025ultra3d} attempted to mitigate this issue by optimizing network computation rather than improving latent compactness.
In this work, our method learns a compact structured latent space directly from native 3D inputs, achieving higher spatial dowmsampling rate and fewer latent tokens.

\paravspace
\paragraph{Large 3D Asset Generation Models and Systems.}
With the rapid expansion of large-scale 3D asset datasets~\cite{deitke2023objaverse,deitke2024objaverse,zhang2025texverse}, there has been a surge in large models and systems capable of generating 3D assets with high-quality shapes and textures~\cite{zhang2024clay,xiang2025structured,chen20243dtopia,hunyuan3d2025hunyuan3d,yang2025pandora3d,li2025step1x,li2025triposg}.
A common paradigm decomposes the generation into two stages: shape generation and multi-view texture synthesis~\cite{zhang2024clay,hunyuan3d2025hunyuan3d,yang2025pandora3d,li2025step1x,li2025step1x,li2025triposg}.
\begin{figure}[t]
	\centering
	\includegraphics[width=\linewidth]{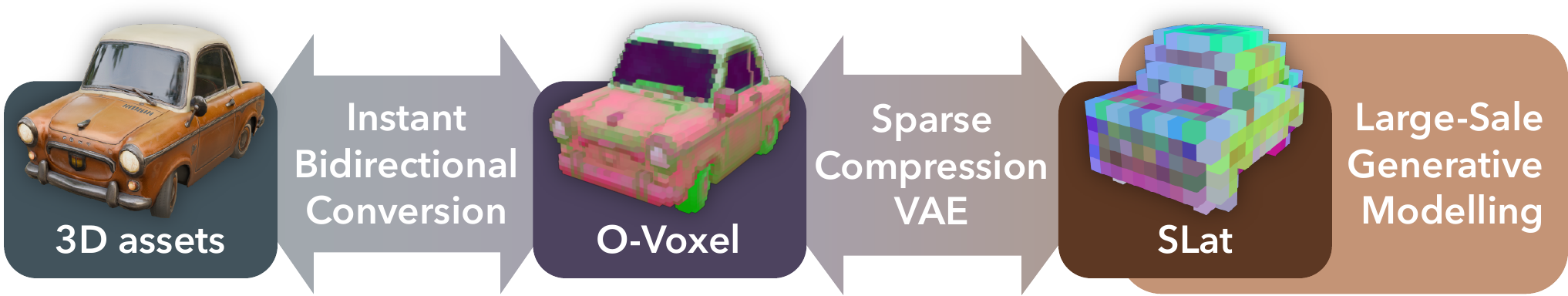}
    \vspace{-20pt}
	\caption{Overview of our approach. We introduce O-Voxel for shape and material representation (Sec.~\ref{sec:N3xus}), based on which we employ Sparse Compression VAEs for compact latent space learning (Sec.~\ref{sec:dcslat}) and large flow models for 3D generation (Sec.~\ref{sec:gen}).}
	\label{fig:overview}
    \vspace{-2pt}
\end{figure}

While this pipeline benefits from powerful 2D image diffusion backbones, it typically requires complex multi-view rendering, baking, and texture alignment, which hinder scalability and often introduce appearance inconsistencies.
\cite{xiang2025structured} tackles 3D generation with material information, yet still relies on multiview baking to merge the generated mesh and 3D Gaussians for asset extraction.
In contrast, our approach performs native, end-to-end 3D asset generation, directly producing high-fidelity, fully textured 3D assets without any view-dependent postprocessing.

\section{Method}\label{sec:methodology}

Our objective is to generate high-resolution 3D assets with arbitrary shape topology and flexible material attributes. An overview of our approach is presented in Fig.~\ref{fig:overview}.

\subsection{O-Voxel: A Native 3D Representation}\label{sec:N3xus}

Given a 3D asset, O-Voxel represents it as a collection of feature tuples associated with \emph{sparse} voxels on a regular 3D grid of resolution $N\times N \times N$:
\begin{equation}
	\boldsymbol{f} = \{(\boldsymbol{f}^{\text{shape}}_i, \boldsymbol{f}^{\text{mat}}_i, \boldsymbol{p}_i)\}_{i=1}^{L},
	\label{eq:N3xus}
\end{equation}
where $\boldsymbol{f}^{\text{shape}}_i$ encodes local geometric information, 
$\boldsymbol{f}^{\text{mat}}_i$ encodes material properties, 
and $\boldsymbol{p}_i \in \{0,1,\ldots,N-1\}^3$ denotes the coordinate of the $i$-th active voxel. Empty voxels that do not intersect with the asset are set inactive.

\begin{figure}[t]
	\centering
	\includegraphics[width=\linewidth]{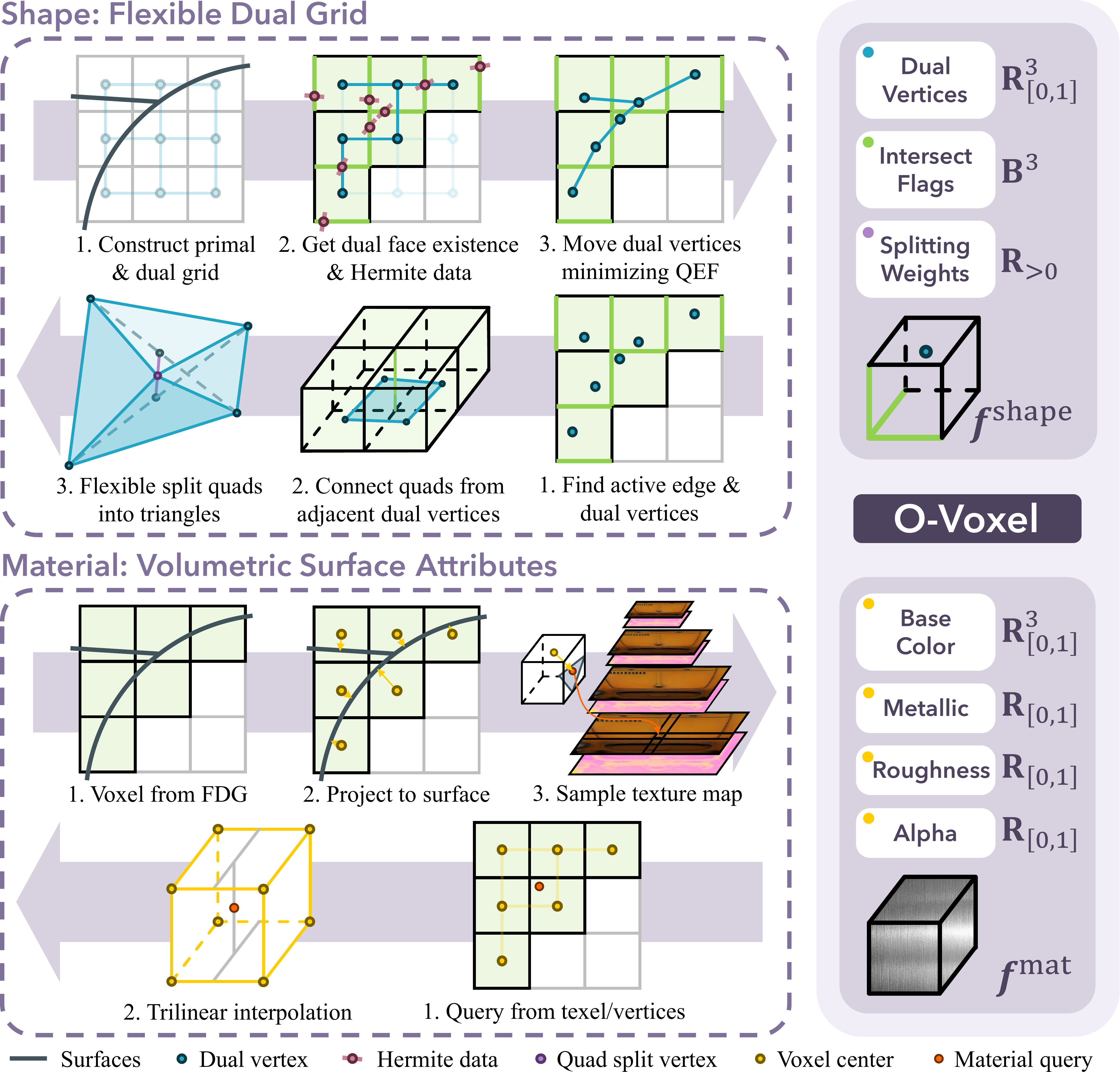}
    \vspace{-17pt}
	\caption{Illustration of O-Voxel and the instant bidirectional convertion between 3D asset and O-Voxel.
    }
	\label{fig:representation}
    \vspace{-2pt}
\end{figure}

\subsubsection{Flexible Dual Grid for Shape}

The O-Voxel can robustly represent surfaces with arbitrary topology, owing to its \emph{Flexible Dual Grid} formulation. In this \emph{dual} grid, one vertex is defined per primal cell and one quadrilateral face per primal edge, connecting dual vertices in adjacent primal cells.
By constructing a dual grid corresponding to the primal regular voxel grid, our algorithm flexibly adjusts the positions of dual vertices and the existence of dual grid faces to accurately represent arbitrary input surface data (illustrated in the first row of Fig.~\ref{fig:representation}).

This formulation is inspired by Dual Contouring (DC)~\cite{ju2002dual,chen2022neural}, an algorithm to extract surfaces from a signed grid with edges tagged by Hermite data (\ie, intersection
points and normals).
The original DC was designed to process descritized scalar fields such as SDFs. Different from DC, we \emph{do not utilize any field representation}. Our approach is straightforward: we directly use the asset’s mesh surface to determine edge intersection flags (rather than detecting sign changes as in DC) and to assign Hermite data. Each edge that intersects the mesh activates the corresponding dual face, and the associated Hermite data are then used to adjust the positions of the dual vertices. Given Hermite data $\{\boldsymbol{q}_i, \boldsymbol{n}_i\}$, we compute the dual vertex $\boldsymbol{v}$ in closed form using the following quadratic error function (QEF):
\begin{equation}\label{eq:qef}
\min_{\boldsymbol{v}\in\text{voxel}} e(\boldsymbol{v}) =
\sum_i d_{\Pi,i}^2
+ \lambda_{\text{bound}} \sum_j d_{L,j}^2
+ \lambda_{\text{reg}}\, d_{\hat{\boldsymbol{q}}}^2.
\end{equation}
The original QEF in DC only contains the first component, which measures the squared distance from $\boldsymbol{v}$ to the plane determined by $\{\boldsymbol{q}_i,\boldsymbol{n}_i\}$, $d_{\Pi,i}^2=(\boldsymbol{n}_i\cdot(\boldsymbol{v}-\boldsymbol{q}_i))^2$. We introduce an additional error term that penalizes the distance between 
$\boldsymbol{v}$ and any boundary edges of the mesh intersecting the primal cell, $d_{L,j}^2 = \left\| (\boldsymbol{v} - \boldsymbol{o}_j) - ((\boldsymbol{v} - \boldsymbol{o}_j) \cdot \boldsymbol{d}_j)\boldsymbol{d}_j \right\|^2$, 
as well as a regularization term that encourages $\boldsymbol{v}$ to stay close 
to the average of the intersecting points, $d_{\hat{\boldsymbol{q}}}^2 = \|\boldsymbol{v} - \bar{\boldsymbol{q}}\|^2$. 
The former guides the dual vertex to align with boundary edges, improving the representation of open surfaces, while the latter encourages a smoother vertex distribution and stabilizes the QEF optimization against singularities.

Based on the above algorithm description, for each active voxel, our geometric feature $\boldsymbol{f}^{\text{shape}}_i$ comprises:
\begin{itemize}[leftmargin=1em]
	\item \textbf{Dual vertex} $\boldsymbol{v}_i \in \mathbb{R}_{[0,1]}^3$, a vertex within the grid to represent  local surface shape.
	\item \textbf{Edge intersection flags} $\boldsymbol{\delta}_i \in \{0,1\}^3$, which determine the quad connections among neighboring dual vertices. We use the surface-edge intersections on 3 predefined voxel edges along the $X$, $Y$, and $Z$ axes, respectively (\eg, those that share the minimum-coordinate corner)\footnote{The flags for the other 9 voxel edges are stored in neighboring voxels.}.
	\item \textbf{Splitting weights} $\gamma_i \in \mathbb{R}_{>0}$, controlling how quadrilateral faces are adaptively subdivided into triangles, following the flexible topology rule in~\cite{shen2023flexicubes}.
\end{itemize}
The conversion algorithm between O-Voxel and mesh is summarized as follows (see also Fig.~\ref{fig:representation} for an illustration).

\textbf{Mesh $\rightarrow$ O-Voxel.}
Given a mesh, we first identify all voxel edges intersecting with the mesh surface and mark their neighboring voxels as active. 
The intersection points and their normals are computed analytically from mesh triangles, yielding Hermite data. The dual vertex for each active voxel is then calculated by solving the QEF in Eq.~\eqref{eq:qef}.

\textbf{O-Voxel $\rightarrow$ Mesh.}
From an O-Voxel, we recover mesh surfaces by connecting dual vertices across intersected edges, forming quadrilateral faces among neighboring active voxels. 
Each quadrilateral can be adaptively subdivided into two triangles guided by the splitting weights, allowing the surface to better conform to local geometric features.

\smallskip
\noindent The Flexible Dual Grid offers several \textbf{key advantages}:
\begin{enumerate}
	\item \emph{Instant bidirectional conversion} — it enables rapid mapping with meshes and supports high-resolution conversion with minimal computation overhead. 
	The costly processes in prior works such as SDF evaluation, flood-fill procedure, and iterative optimization are not needed.
	\item \emph{Arability-topology modeling} — it is free from the watertight and manifold constraints, enabling robust handling of arbitrary geometry including self-intersecting surfaces and fully-enclosed interior structures. 
	\item \emph{High precision and sharp feature preservation} — the dual vertices are aligned with local geometric features by algorithm design~\cite{ju2002dual}, allowing for the preservation of sharp features. Additionally, the dual vertex positions and splitting weights can receive learnable adjustments by neural network using other supervisions such as rendering loss (see Section~\ref{sec:vae_training}), further enhancing the flexibility and precision of the geometry. 
\end{enumerate}

\subsubsection{Volumetric Attributes for Material}

The O-Voxel can model arbitrary surface attributes aligned with surface geometry, including color and other material attributes. In this work, we implement physically-based rendering (PBR) parameters to capture the intrinsic light–surface interaction characteristics of the materials. Specifically, our material feature $\boldsymbol{f}^{\text{mat}}_i$ for each active voxel consists of six channels:
\begin{equation}
	\boldsymbol{f}^{\text{mat}}_i = (\boldsymbol{c}_i, m_i, r_i, \alpha_i),
\end{equation}
where $\boldsymbol{c}_i \in \mathbb{R}_{[0,1]}^3$ denotes the \emph{base color}, 
$m_i \in \mathbb{R}_{[0,1]}$ the \emph{metallic ratio}, 
$r_i \in \mathbb{R}_{[0,1]}$ the \emph{roughness}, 
and $\alpha_i \in \mathbb{R}_{[0,1]}$ the \emph{opacity}.
This parameterization follows the standard PBR convention widely adopted in modern physically-based rendering pipelines.
The conversion between O-Voxel data and mesh texture is simple and fast, as illustrated in Fig.~\ref{fig:representation}.

\textbf{Texture $\rightarrow$ O-Voxel.}
For an active voxel, we project its center onto each intersected triangle and sample each material attribute from the texture map using UV coordinates and appropriate mipmap levels. The sampled attributes are weighted-averaged based on the point-to-surface distances to obtain the final value.  More details of this process can be found in the \emph{supplementary material}.

\textbf{O-Voxel $\rightarrow$ Texture.}
During reconstruction, for each query point -- either a mesh vertex position for vertex coloring or a 3D surface point corresponding to a texture map texel -- material attributes are obtained via trilinear interpolation of the neighboring voxel attributes. The reconstructed mesh is then ready for rendering without need for any additional post-processing.

\subsection{Sparse Compression VAE}\label{sec:dcslat}

We apply a VAE to learn a proper latent space from O-Voxel data. Our goal is to obtain a compact latent space that facilitates efficient, high-resolution 3D generation. We design a Sparse Compression VAE (SC-VAE) to achieve this.

\subsubsection{Network Architecture}
\begin{figure}[t]
	\centering
	\includegraphics[width=\linewidth]{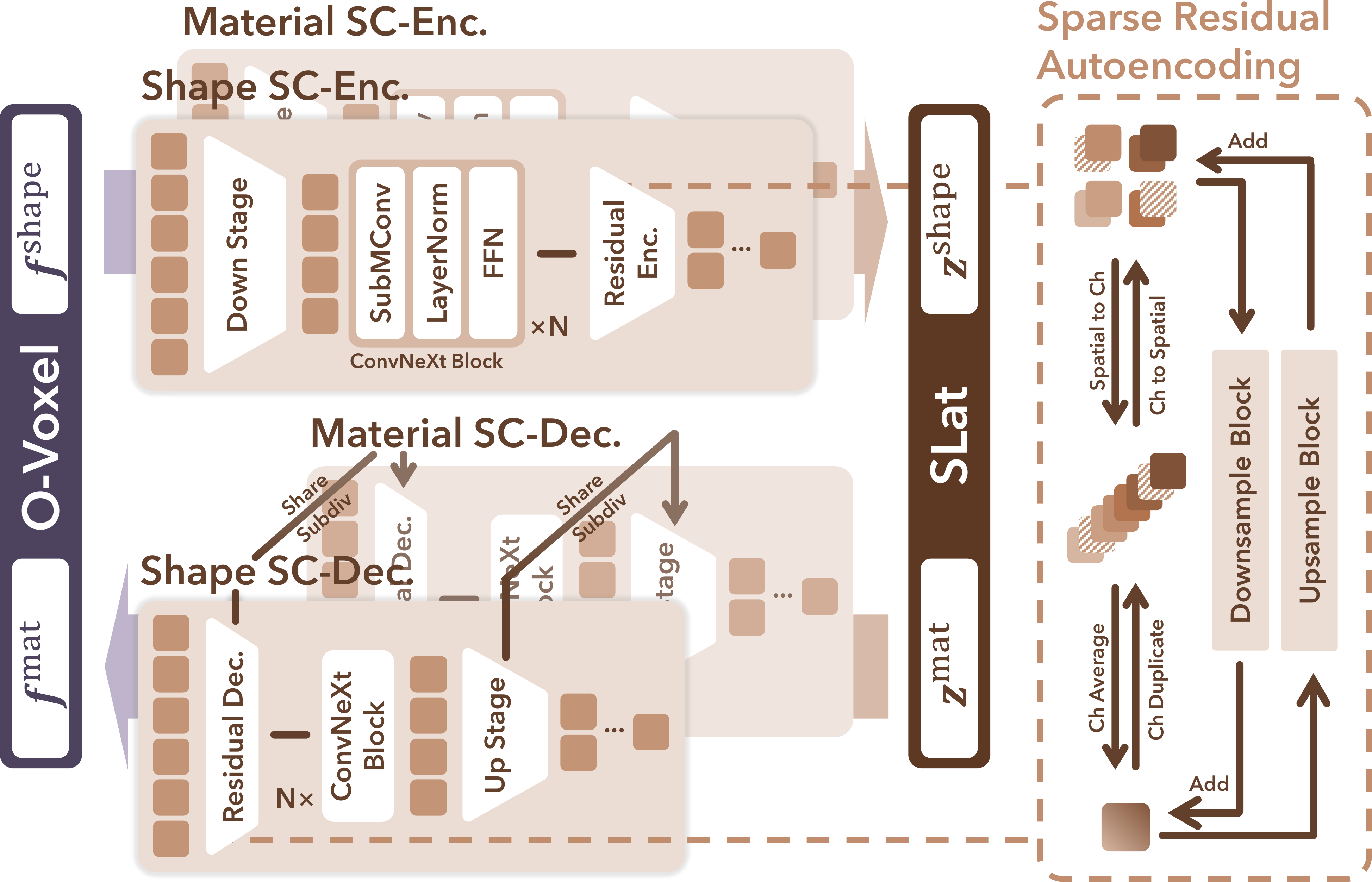}
    \vspace{-18pt}
	\caption{The network structure of SC-VAE.
    }
	\label{fig:vae}
    \vspace{-10pt}
\end{figure}

The architecture of our SC-VAE is illustrated in Fig.~\ref{fig:vae}.
Unlike transformer-based designs in priors work~\cite{xiang2025structured, he2025sparseflex}, our SC-VAE employs a \emph{fully sparse-convolutional network} that is both computationally efficient at high resolutions and generalizes well across scales.
Following a U-shaped VAE design~\cite{rombach2022high}, our encoder hierarchically downsamples sparse voxel features through multiple residual blocks, and the decoder mirrors this process for reconstruction.
We meticulously design the residual and (down/up)sampling blocks to enable high-compression encoding and faithful recovery.

\paravspace
\paragraph{Sparse Residual Autoencoding Layer.}
We adapt the Residual Autoencoding principle from DC-AE~\cite{chen2024deep} to sparse voxel data by introducing non-parametric residual shortcuts within downsampling and upsampling blocks.
These shortcuts mitigate optimization challenges under high spatial compression by rearranging information between space and channel dimensions in the sparse grid.

Specifically, for a downsampling factor of 2, we aggregate each voxel’s eight children into its channel dimension.
Given input features $F_{\text{fine}}\in\mathbb{R}^{C}$ and target coarse features $F_{\text{coarse}}\in\mathbb{R}^{C'}$ (typically $C'=2C$), we have:
\begin{equation}
	\begin{split}
		\quad F_{\text{coarse}}^{\text{raw}} &= \operatorname{stack}\big(F_{\text{child}_1},\dots,F_{\text{child}_8}\big)\in\mathbb{R}^{8C},\\
		F_{\text{coarse}} &= \operatorname{avg\_groups}\big(F_{\text{coarse}}^{\text{raw}}\big)\in\mathbb{R}^{C'},
	\end{split}
\end{equation}
where the $\operatorname{avg\_groups}$ operation averages grouped channels to produce a coarse residual estimate. Missing voxels contribute zero vectors due to sparsity.

During upsampling, a symmetric channel-to-space shortcut distributes each coarse feature back to its neighborhood:
\begin{equation}
	\begin{split}
		\quad F_{\text{fine}}^{\text{raw}} &= \operatorname{unstack}\big(F_{\text{coarse}}\big)\in\mathbb{R}^{8C'/8},\\
		F_{\text{fine}} &= \operatorname{dup\_groups}\big(F_{\text{fine}}^{\text{raw}}\big)\in\mathbb{R}^{C},
	\end{split}
\end{equation}
where $\operatorname{dup\_groups}$ copies channels within each group to match the target dimension.

\paravspace
\paragraph{Early-pruning Upsampler.}
To further enhance efficiency, we employ an \emph{early-pruning} mechanism~\cite{ren2024xcube} for the upsampler.
Before each upsampling step, the module predicts a binary mask $\boldsymbol{\hat{\rho}}\in\{0,1\}^8$ specifying the active child voxels of each parent node.
Inactive nodes are skipped subsequently, thus greatly reducing runtime and memory cost.

\paravspace
\paragraph{Optimized Residual Block.}
Sparse convolutions exhibit low effective computation and parameter efficiency for high sparsity data. In light of this, we redesign the residual block by reducing convolutional layers and incorporating point-wise MLPs for richer feature transformation. Specifically, we substitute the standard design of two conv layers with a single conv layer and fewer normalization and activation layers, following the ConvNeXt-style~\cite{liu2022convnet} simplification. The second conv is replaced by a wide point-wise MLP -- analogous to a Transformer FFN -- which expands  channel dimensions for enhanced nonlinearity and representation. This modification does not affect efficiency but improves reconstruction quality, as demonstrated in our experiments.

\subsubsection{VAE Training}\label{sec:vae_training}

SC-VAE is trained in two stages.
In the first stage, we use low-resolution data to quickly stabilize learning with direct O-Voxel reconstruction loss and KL loss. For geometry features, Mean-Squared-Error (MSE) and Binary Cross Entropy (BCE) losses are applied on dual vertex positions $\boldsymbol{v}$ and dual face flags $\boldsymbol{\delta}$, respectively.
Material attributes $\boldsymbol{f}^{\text{mat}}$ and the pruning mask $\boldsymbol{\rho}$ are supervised by the L1 and BCE loss, receptively:
\begin{equation}
	\begin{split}
		\mathcal{L}_{\text{s1}} &= \lambda_{\text{v}}|\hat{\boldsymbol{v}}-\boldsymbol{v}|_2^2 + \lambda_{\delta}\operatorname{BCE}(\hat{\boldsymbol{\delta}},\boldsymbol{\delta}) + \lambda_{\boldsymbol{\rho}}\operatorname{BCE}(\hat{\boldsymbol{\rho}},\boldsymbol{\rho}) \\
		&+ \lambda_{\text{mat}}|\hat{\boldsymbol{f}}^{\text{mat}} - \boldsymbol{f}^{\text{mat}}|_1 + \lambda_{\text{KL}}\mathcal{L}_{\text{KL}}.
	\end{split}
\end{equation}

\vspace{-3px}
In the second stage, we add rendering-based perceptual supervision at high resolution to enhance geometric and material fidelity.
We render \emph{mask}, \emph{depth}, and \emph{normal} maps and supervise them with L1 loss, augmented with SSIM and LPIPS terms on normals.
The material attributes are rendered and supervised by these perceptual losses as well. The loss can be written as:
\begin{equation}
	\mathcal{L}_{\text{s2}} = \mathcal{L}_{\text{s1}} + \mathcal{L}_{\text{render}}.
\end{equation}
We randomly place cameras around with a shallow near plane to \emph{slice through the surface}, encouraging the model to capture both external and internal structures.
More details of the losses are provided in the \emph{supplementary material.}

To facilitate a sequential generation scheme for shape and material (in particular, to enable the application of material generation for given shapes), we learn decoupled latent spaces with two SC-VAEs: one models shape, while the other models material conditioned on the shape VAE's subdivision structures during upsampling.

\subsection{Generative Modeling}\label{sec:gen}

Built upon the learned latent space, we construct a scalable generative framework following the overall design of \cite{xiang2025structured}.
We adopt full DiT-based architectures~\cite{peebles2023scalable} trained with the flow matching paradigm~\cite{lipman2023flow} and extend the pipeline of \cite{xiang2025structured} to fully leverage the power of our new latents.

\paravspace
\paragraph{Model and Generation Pipeline.}
The complete generation process unfolds in three stages with three models:
1) \emph{sparse structure generation}, which predicts the occupancy layout of the sparse voxel grid;
2) \emph{geometry generation}, which produces geometry latents within active voxels; and
3) \emph{material generation}, which synthesizes material latents aligned to the geometry structure.
The first two stages largely follow the strategy of \cite{xiang2025structured}, forming the geometric backbone of the asset.
The novel \emph{material generation} stage models PBR materials directly in the native 3D space.
A sparse DiT predicts material latents conditioned jointly on the input image and the generated geometry latents.
This design unifies geometric and material generation in the same native 3D latent domain and ensures their spatially alignment under arbitrary topology.

\paravspace
\paragraph{Architectural and Training Details.}
All our DiT modules employ the AdaLN-single modulation~\cite{chen2024pixart} and Rotary Position Embedding (RoPE)~\cite{su2024roformer} for better scalability and cross-resolution generalization.
Image conditioning features are extracted from DINOv3-L~\cite{simeoni2025dinov3}.
Benefiting from the high spatial compression achieved by SC-VAE, our sparse DiTs discard the convolutional packing and skip connection designs in \cite{xiang2025structured}, resulting in a vanilla-style DiT which reduces complexity and improves efficiency.

We first train the sparse structure generation with $512\times 512$ conditioning images to learn coarse occupancy priors and establish the global sparse layout.
In the subsequent stages, training proceeds in a progressive manner, gradually increasing both the spatial and visual resolution.
The geometry and material generators are scaled from $512^3$ outputs ($32^3$ latent resolution) to $1024^3$ outputs ($64^3$ latent resolution), with the conditioning image resolution correspondingly increased to $1024$.
This progressive strategy allows the learned priors to transfer smoothly across resolutions, enabling efficient training of large-scale sparse DiTs while maintaining fidelity in both geometry and material.

\begin{figure*}[ht]
	\centering
	\includegraphics[width=\linewidth]{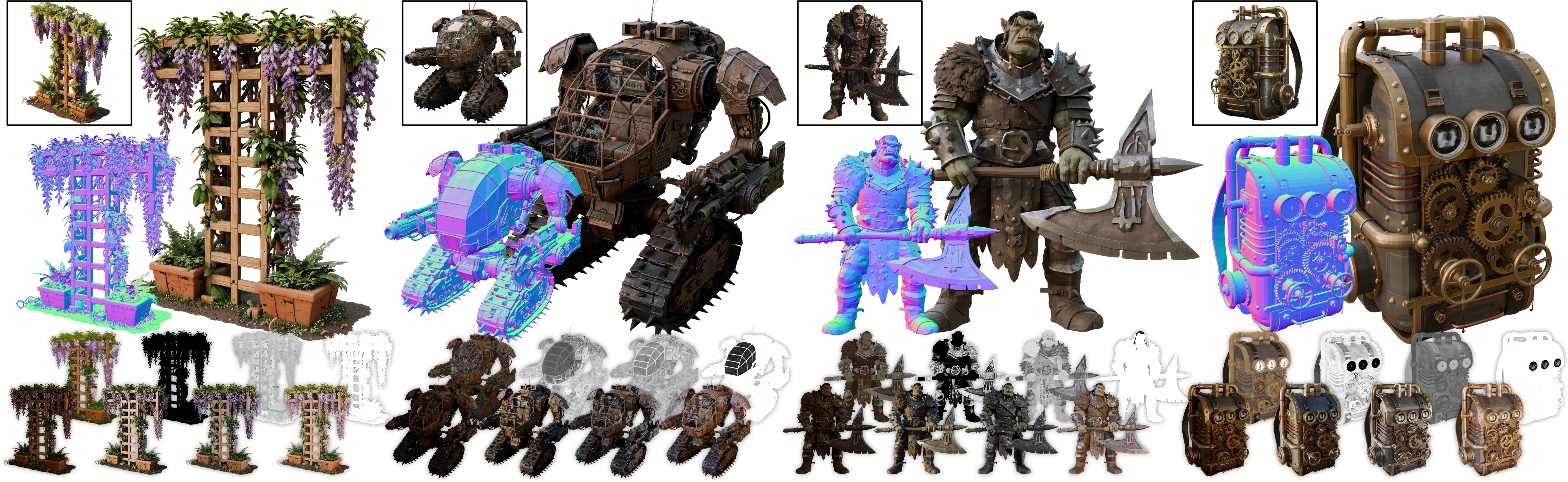}
    \vspace{-20pt}
	\caption{High-quality 3D assets generated by our method, featuring intricate geometric details and physically accurate materials with high visual fidelity, including thin structures, open surfaces, and translucent regions that highlight the model’s expressive capability.}
	\label{fig:results}
    \vspace{-8pt}
\end{figure*}

\begin{figure*}[ht]
	\centering
	\includegraphics[width=\linewidth]{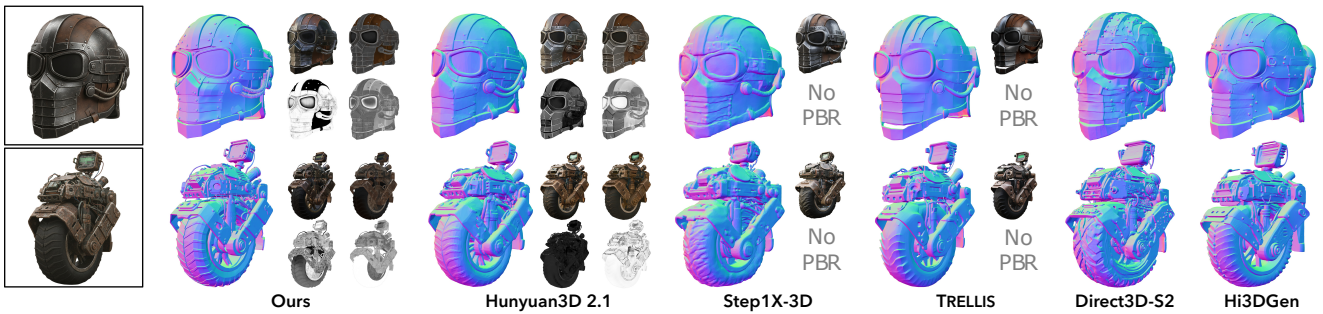}
    \vspace{-20pt}
	\caption{Visual comparison showing the normal (main image) and the final render, base color, metallic, and roughness (small images).}
	\label{fig:comparison}
    \vspace{-8pt}
\end{figure*}

\section{Experiments}

\paragraph{Implementation details.}
Our SC-VAE is trained using the Trellis-500K setup~\cite{xiang2025structured} after filtering out assets without PBR materials, resulting in a curated collection from Objaverse-XL~\cite{deitke2024objaverse}, ABO~\cite{collins2022abo}, and HSSD~\cite{khanna2023hssd}.
We use an optimized Triton~\cite{tillet2019triton} implementation for Submanifold convolution~\cite{graham2017submanifold} to further improve training speed. SC-VAEs are trained on 16 H100 GPUs with a batch size of 128.

The generative model is trained on an extended collection of about 800K assets, augmented with TexVerse~\cite{zhang2025texverse} to enrich PBR diversity and realism.
For image prompts, we render 16 views per asset in Blender~\cite{blender} with randomized FoVs and lighting.
All models are trained using AdamW~\cite{loshchilov2017decoupled} (learning rate $1\times10^{-4}$, weight decay $0.01$) with classifier-free guidance (drop rate $0.1$).
Each DiT in our framework contains approximately 1.3B parameters (width: 1536, blocks: 30, heads: 12, MLP width: 8192),
trained on 32 H100 GPUs with a batch size of 256.

For reconstruction evaluation, we use the Toys4K benchmark~\cite{stojanov2021using} together with a curated test set containing 90 assets featuring complex PBR materials and detailed shapes from recent Sketchfab assets~\cite{sketchfab2025} released within the past two years. Both test sets are unseen during training.
For generation quality comparison and user studies, 100 AI-generated image prompts~\cite{fortin2025nanobanana} are used to ensure training–testing disjointness.
Split-sum renderer from nvdiffrec~\cite{munkberg2022extracting} is used for PBR asset rendering. All runtime statistics are reported on an NVIDIA A100 GPU.
Additional details are provided in the \emph{supplementary material}.

\begin{table*}[t]
        \centering
        \scriptsize
        \caption{Comparison of shape reconstruction efficiency and fidelity. MD and CD is reported $\times 10^6$;  runtime meaured on an A100 GPU.}
        \vspace{-8pt}
        \renewcommand{\arraystretch}{1}
        \setlength{\tabcolsep}{3pt}
        \begin{tabular}{cr@{\,}lcccccccccccccc}
        \toprule
        & & & & & \multicolumn{6}{c}{\textbf{Toys4K}} & \multicolumn{6}{c}{\textbf{Sketchfab Featured}} \\
        [-0.5ex]\cmidrule(lr){6-11}\cmidrule(lr){12-17}
        \textbf{Method} & \multicolumn{2}{c}{\textbf{\#Token \tiny{(\#Dim)}}} & $\boldsymbol{f_{\text{down}}}$ & \textbf{Dec. {\tiny(s)}}$\downarrow$ & \multicolumn{2}{c}{\textbf{All Surface}} & \multicolumn{2}{c}{\textbf{Outer Surface}} & \multicolumn{2}{c}{\textbf{Normal}} & \multicolumn{2}{c}{\textbf{All Surface}} & \multicolumn{2}{c}{\textbf{Outer Surface}} & \multicolumn{2}{c}{\textbf{Normal}} \\
        [-0.5ex]\cmidrule(lr){6-7}\cmidrule(lr){8-9}\cmidrule(lr){10-11}\cmidrule(lr){12-13}\cmidrule(lr){14-15}\cmidrule(lr){16-17}
         & & & & & \textbf{MD$\downarrow$} & \textbf{F1\textsubscript{1e-8}$\uparrow$} & \textbf{CD$\downarrow$} & \textbf{F1\textsubscript{1e-6}$\uparrow$} & \textbf{PSNR$\uparrow$} & \textbf{LPIPS$\downarrow$} & \textbf{MD$\downarrow$} & \textbf{F1\textsubscript{1e-8}$\uparrow$} & \textbf{CD$\downarrow$} & \textbf{F1\textsubscript{1e-6}$\uparrow$} & \textbf{PSNR$\uparrow$} & \textbf{LPIPS$\downarrow$} \\
        \midrule
        Dora & 2.0K & \tiny{(131K)} & -- & \cellcolor[rgb]{1.00,0.81,0.80}37.7 & \cellcolor[rgb]{1.00,0.80,0.80}366.1 & \cellcolor[rgb]{1.00,0.81,0.80}0.019 & \cellcolor[rgb]{1.00,0.80,0.80}325.3 & \cellcolor[rgb]{1.00,0.86,0.80}0.140 & \cellcolor[rgb]{1.00,0.82,0.80}27.26 & \cellcolor[rgb]{1.00,0.86,0.80}0.117 & \cellcolor[rgb]{1.00,0.81,0.80}987.2 & \cellcolor[rgb]{1.00,0.81,0.80}0.020 & \cellcolor[rgb]{1.00,0.81,0.80}1182. & \cellcolor[rgb]{1.00,0.86,0.80}0.130 & \cellcolor[rgb]{1.00,0.80,0.80}22.02 & \cellcolor[rgb]{1.00,0.80,0.80}0.175 \\
        Dora & 4.1K & \tiny{(262K)} & -- & \cellcolor[rgb]{1.00,0.80,0.80}43.0 & \cellcolor[rgb]{1.00,0.80,0.80}360.8 & \cellcolor[rgb]{1.00,0.81,0.80}0.019 & \cellcolor[rgb]{1.00,0.80,0.80}327.3 & \cellcolor[rgb]{1.00,0.86,0.80}0.138 & \cellcolor[rgb]{1.00,0.82,0.80}27.32 & \cellcolor[rgb]{1.00,0.87,0.80}0.115 & \cellcolor[rgb]{1.00,0.80,0.80}1139. & \cellcolor[rgb]{1.00,0.81,0.80}0.020 & \cellcolor[rgb]{1.00,0.80,0.80}1351. & \cellcolor[rgb]{1.00,0.86,0.80}0.130 & \cellcolor[rgb]{1.00,0.80,0.80}22.09 & \cellcolor[rgb]{1.00,0.80,0.80}0.174 \\
        \textsc{Trellis} & 9.6K & \tiny{(77K)} & 4$\times$ & \underline{\cellcolor[rgb]{0.82,1.00,0.80}0.108} & \cellcolor[rgb]{1.00,0.85,0.80}85.07 & \cellcolor[rgb]{1.00,0.83,0.80}0.074 & \cellcolor[rgb]{0.90,1.00,0.80}2.755 & \cellcolor[rgb]{0.95,1.00,0.80}0.544 & \cellcolor[rgb]{1.00,0.89,0.80}30.29 & \cellcolor[rgb]{0.99,1.00,0.80}0.067 & \cellcolor[rgb]{1.00,0.91,0.80}49.20 & \cellcolor[rgb]{1.00,0.81,0.80}0.031 & \cellcolor[rgb]{0.93,1.00,0.80}8.171 & \cellcolor[rgb]{1.00,0.93,0.80}0.286 & \cellcolor[rgb]{1.00,0.87,0.80}24.31 & \cellcolor[rgb]{1.00,0.96,0.80}0.110 \\
        Direct3D-S2 512 & 3.0K & \tiny{(48K)} & 8$\times$ & \cellcolor[rgb]{1.00,1.00,0.80}1.86 & \cellcolor[rgb]{1.00,0.83,0.80}137.6 & \cellcolor[rgb]{1.00,0.80,0.80}0.002 & \cellcolor[rgb]{1.00,0.96,0.80}23.85 & \cellcolor[rgb]{1.00,0.81,0.80}0.027 & \cellcolor[rgb]{1.00,0.80,0.80}26.34 & \cellcolor[rgb]{1.00,0.80,0.80}0.138 & \cellcolor[rgb]{1.00,0.88,0.80}128.0 & \cellcolor[rgb]{1.00,0.80,0.80}0.002 & \cellcolor[rgb]{0.99,1.00,0.80}26.64 & \cellcolor[rgb]{1.00,0.81,0.80}0.032 & \cellcolor[rgb]{1.00,0.82,0.80}22.76 & \cellcolor[rgb]{1.00,0.84,0.80}0.159 \\
        Direct3D-S2 1024 & 17K & \tiny{(271K)} & 8$\times$ & \cellcolor[rgb]{1.00,0.88,0.80}13.0 & \cellcolor[rgb]{1.00,0.86,0.80}73.17 & \cellcolor[rgb]{1.00,0.80,0.80}0.001 & \cellcolor[rgb]{1.00,1.00,0.80}13.45 & \cellcolor[rgb]{1.00,0.80,0.80}0.001 & \cellcolor[rgb]{1.00,0.82,0.80}27.38 & \cellcolor[rgb]{1.00,0.81,0.80}0.134 & \cellcolor[rgb]{1.00,0.90,0.80}70.13 & \cellcolor[rgb]{1.00,0.80,0.80}0.001 & \cellcolor[rgb]{0.96,1.00,0.80}14.06 & \cellcolor[rgb]{1.00,0.80,0.80}0.000 & \cellcolor[rgb]{1.00,0.85,0.80}23.82 & \cellcolor[rgb]{1.00,0.89,0.80}0.138 \\
        SparseFlex 512 & 56K & \tiny{(452K)} & 4$\times$ & \cellcolor[rgb]{0.98,1.00,0.80}1.22 & \cellcolor[rgb]{1.00,1.00,0.80}1.303 & \cellcolor[rgb]{0.90,1.00,0.80}0.735 & \cellcolor[rgb]{0.82,1.00,0.80}0.8366 & \cellcolor[rgb]{0.81,1.00,0.80}0.840 & \cellcolor[rgb]{0.96,1.00,0.80}36.56 & \cellcolor[rgb]{0.87,1.00,0.80}0.027 & \cellcolor[rgb]{0.99,1.00,0.80}3.221 & \cellcolor[rgb]{0.98,1.00,0.80}0.487 & \cellcolor[rgb]{0.86,1.00,0.80}1.881 & \cellcolor[rgb]{0.83,1.00,0.80}0.783 & \cellcolor[rgb]{0.98,1.00,0.80}29.46 & \cellcolor[rgb]{0.90,1.00,0.80}0.055 \\
        SparseFlex 1024 & 225K & \tiny{(1799K)} & 4$\times$ & \cellcolor[rgb]{1.00,0.96,0.80}3.21 & \cellcolor[rgb]{0.95,1.00,0.80}0.3132 & \cellcolor[rgb]{0.85,1.00,0.80}0.845 & \cellcolor[rgb]{0.82,1.00,0.80}0.8062 & \cellcolor[rgb]{0.81,1.00,0.80}0.843 & \cellcolor[rgb]{0.94,1.00,0.80}37.34 & \cellcolor[rgb]{0.91,1.00,0.80}0.042 & \cellcolor[rgb]{0.94,1.00,0.80}0.7593 & \underline{\cellcolor[rgb]{0.88,1.00,0.80}0.684} & \cellcolor[rgb]{0.83,1.00,0.80}1.129 & \cellcolor[rgb]{0.81,1.00,0.80}0.828 & \underline{\cellcolor[rgb]{0.89,1.00,0.80}32.12} & \cellcolor[rgb]{0.86,1.00,0.80}0.036 \\
        \textit{Ours 512} & 2.2K & \tiny{(70K)} & 16$\times$ & \textbf{\cellcolor[rgb]{0.80,1.00,0.80}0.077} & \underline{\cellcolor[rgb]{0.87,1.00,0.80}0.0323} & \underline{\cellcolor[rgb]{0.83,1.00,0.80}0.888} & \underline{\cellcolor[rgb]{0.80,1.00,0.80}0.5890} & \underline{\cellcolor[rgb]{0.80,1.00,0.80}0.851} & \underline{\cellcolor[rgb]{0.88,1.00,0.80}39.54} & \underline{\cellcolor[rgb]{0.82,1.00,0.80}0.013} & \underline{\cellcolor[rgb]{0.88,1.00,0.80}0.1572} & \cellcolor[rgb]{0.92,1.00,0.80}0.613 & \underline{\cellcolor[rgb]{0.81,1.00,0.80}0.7447} & \underline{\cellcolor[rgb]{0.81,1.00,0.80}0.833} & \cellcolor[rgb]{0.93,1.00,0.80}31.00 & \underline{\cellcolor[rgb]{0.85,1.00,0.80}0.034} \\
        \textit{Ours 1024} & 9.6K & \tiny{(306K)} & 16$\times$ & \cellcolor[rgb]{0.89,1.00,0.80}0.301 & \textbf{\cellcolor[rgb]{0.80,1.00,0.80}0.0042} & \textbf{\cellcolor[rgb]{0.80,1.00,0.80}0.971} & \textbf{\cellcolor[rgb]{0.80,1.00,0.80}0.5660} & \textbf{\cellcolor[rgb]{0.80,1.00,0.80}0.855} & \textbf{\cellcolor[rgb]{0.80,1.00,0.80}43.11} & \textbf{\cellcolor[rgb]{0.80,1.00,0.80}0.005} & \textbf{\cellcolor[rgb]{0.80,1.00,0.80}0.0170} & \textbf{\cellcolor[rgb]{0.80,1.00,0.80}0.868} & \textbf{\cellcolor[rgb]{0.80,1.00,0.80}0.6402} & \textbf{\cellcolor[rgb]{0.80,1.00,0.80}0.850} & \textbf{\cellcolor[rgb]{0.80,1.00,0.80}35.26} & \textbf{\cellcolor[rgb]{0.80,1.00,0.80}0.013} \\
        \bottomrule
        \end{tabular}
        \vspace{-8pt}
        \label{tab:reconstruction}
\end{table*}

\subsection{3D Asset Reconstruction}

\paragraph{Shape Reconstruction.}
We compare with four representative baselines: Dora~\cite{chen2025dora} based on Shape2Vecset; \textsc{Trellis}~\cite{xiang2025structured}, Direct3D-s2~\cite{wu2025direct3d}, and SparseFlex~\cite{he2025sparseflex} based on sparse voxel structure.
For evaluation, we employ multiple metrics: (i) Mesh Distance (MD) calculated as Bidirectional Point-to-Mesh Distance with F1-score to measure reconstruction fidelity of meshes including internal structures; (ii) Chamfer Distance with F1-score computed on point clouds sampled from visible surfaces, focusing only on external shapes; and (iii) surface quality metrics using PSNR and LPIPS of rendered normal maps.

As shown in Table~\ref{tab:reconstruction}, our method consistently outperforms all baselines by a substantial margin across every metric, despite using only a modest number of tokens and requiring significantly less runtime. This demonstrates not only its superior geometric fidelity, finer detail preservation, and more accurate modeling of internal structures, but also the high efficiency of our approach.

\paravspace
\paragraph{Material Reconstruction.}
As no suitable baseline exists for encoding only material properties given shapes, we report metrics solely for our method. We assess the fidelity of directly rendered PBR attribute maps and shaded images using PSNR and LPIPS. Our method achieves 38.89 dB / 0.033 on PBR attributes and 38.69 dB / 0.026 on shaded images, demonstrating faithful material reproduction and consistent geometry–appearance alignment.

\begin{table}[t]
        \centering
        \scriptsize
        \caption{Comparison of image-to-3D generation results. -N: measured with normal map.}
        \vspace{-8pt}
        \renewcommand{\arraystretch}{1}
        \setlength{\tabcolsep}{2pt}
        \begin{tabular}{ccccccc}
        \toprule
        \multirow{2}{*}{\textbf{Method}} & \multicolumn{4}{c}{\textbf{Alignment}} & \multicolumn{2}{c}{\textbf{Quality {\tiny(User Study)}}} \\
        [-0.5ex]\cmidrule(lr){2-5}\cmidrule(lr){6-7}
         & \textbf{CLIP$\uparrow$} & \textbf{CLIP{\tiny-N}$\uparrow$} & \textbf{ULIP-2$\uparrow$} & \textbf{Uni3D$\uparrow$} & \textbf{Pref\%$\uparrow$} & \textbf{Pref{\tiny-N}\%$\uparrow$} \\
        \midrule
        \textsc{Trellis} & \underline{\cellcolor[rgb]{1.00,0.91,0.80}0.876} & \cellcolor[rgb]{1.00,1.00,0.80}0.748 & \cellcolor[rgb]{0.84,1.00,0.80}0.470 & \cellcolor[rgb]{0.94,1.00,0.80}0.414 & \cellcolor[rgb]{1.00,0.80,0.80}6.40\% & \cellcolor[rgb]{1.00,0.81,0.80}2.82\% \\
        Hi3DGen & -- & \cellcolor[rgb]{0.90,1.00,0.80}0.753 & \cellcolor[rgb]{1.00,0.80,0.80}0.395 & \cellcolor[rgb]{1.00,0.80,0.80}0.373 & -- & \cellcolor[rgb]{1.00,0.84,0.80}6.57\% \\
        Direct3D-S2 & -- & \cellcolor[rgb]{1.00,0.96,0.80}0.746 & \cellcolor[rgb]{1.00,0.92,0.80}0.420 & \cellcolor[rgb]{1.00,0.92,0.80}0.392 & -- & \underline{\cellcolor[rgb]{1.00,0.87,0.80}12.2\%} \\
        Step1X-3D & \cellcolor[rgb]{1.00,0.89,0.80}0.875 & \cellcolor[rgb]{1.00,0.80,0.80}0.738 & \cellcolor[rgb]{0.87,1.00,0.80}0.464 & \cellcolor[rgb]{0.96,1.00,0.80}0.411 & \cellcolor[rgb]{1.00,0.84,0.80}11.8\% & \cellcolor[rgb]{1.00,0.80,0.80}0.469\% \\
        Hunyuan3D 2.1 & \cellcolor[rgb]{1.00,0.80,0.80}0.869 & \underline{\cellcolor[rgb]{0.89,1.00,0.80}0.753} & \underline{\cellcolor[rgb]{0.81,1.00,0.80}0.474} & \underline{\cellcolor[rgb]{0.86,1.00,0.80}0.427} & \underline{\cellcolor[rgb]{1.00,0.85,0.80}13.3\%} & \cellcolor[rgb]{1.00,0.84,0.80}7.51\% \\
        \textit{Ours} & \textbf{\cellcolor[rgb]{0.80,1.00,0.80}0.894} & \textbf{\cellcolor[rgb]{0.80,1.00,0.80}0.758} & \textbf{\cellcolor[rgb]{0.80,1.00,0.80}0.477} & \textbf{\cellcolor[rgb]{0.80,1.00,0.80}0.436} & \textbf{\cellcolor[rgb]{0.80,1.00,0.80}66.5\%} & \textbf{\cellcolor[rgb]{0.80,1.00,0.80}69.0\%} \\
        \bottomrule
        \end{tabular}
        \vspace{-8pt}
        \label{tab:generation}
\end{table}

\subsection{Image to 3D Generation}

We next evaluate the generative capabilities of our framework by producing 3D assets conditioned on input images. Fig.~\ref{fig:results} presents representative results, illustrating both geometric fidelity and material realism.

Leveraging the compact latent space compressing O-Voxels, our method generates assets that faithfully preserve fine-scale structures, sharp surface features, and internally complex or non-manifold shapes—ranging from detailed gears, enclosed cockpit and open leaves and flowers.
It also reproduces vivid, realistic PBR textures with physically consistent shading under novel lighting, including challenging translucent or reflective materials such as glass and metal.
Together, These results demonstrate that our native 3D latent space enables the generation of tightly aligned, high-fidelity geometry and photorealistic appearance, even for topologically complex assets.

\paravspace
\paragraph{Qualitative comparison.}
We compare our approach against state-of-the-art 3D generation systems, including \textsc{Trellis}~\cite{xiang2025structured}, Hi3DGen~\cite{ye2025hi3dgen}, Direct3D-s2~\cite{wu2025direct3d}, Step1X-3D~\cite{li2025step1x}, and Hunyuan3D 2.1~\cite{hunyuan3d2025hunyuan3d}. Samples from Fig.~\ref{fig:comparison} demonstrate that our method achieves superior generation quality—delivering accurate and detailed geometry, physically plausible materials, and faithful prompt alignment.

\paravspace
\paragraph{Quantitative comparison.}
We conduct quantitative evaluation using AI-generated images. Visual alignment is measured with the CLIP score~\cite{radford2021learning}, while multimodal models ULIP-2~\cite{xue2024ulip} and Uni3D~\cite{zhou2023uni3d} are employed to assess the geometric similarity. Table~\ref{tab:generation} shows that our method achieves the highest alignment score across all metrics, demonstrating clear superiority in visual and geometric consistency.

We also conduct a user study with about 40 participants to evaluate perceptual quality.
Using 100 AI-generated image prompts, we generate assets with each method under identical conditions without curation and collect preference votes.
Table~\ref{tab:generation} shows our method is favored by participants, highlighting its clear superiority in visual realism, richness of geometric detail, and alignment with input prompts.

\begin{figure}[ht]
	\centering
	\includegraphics[width=\linewidth]{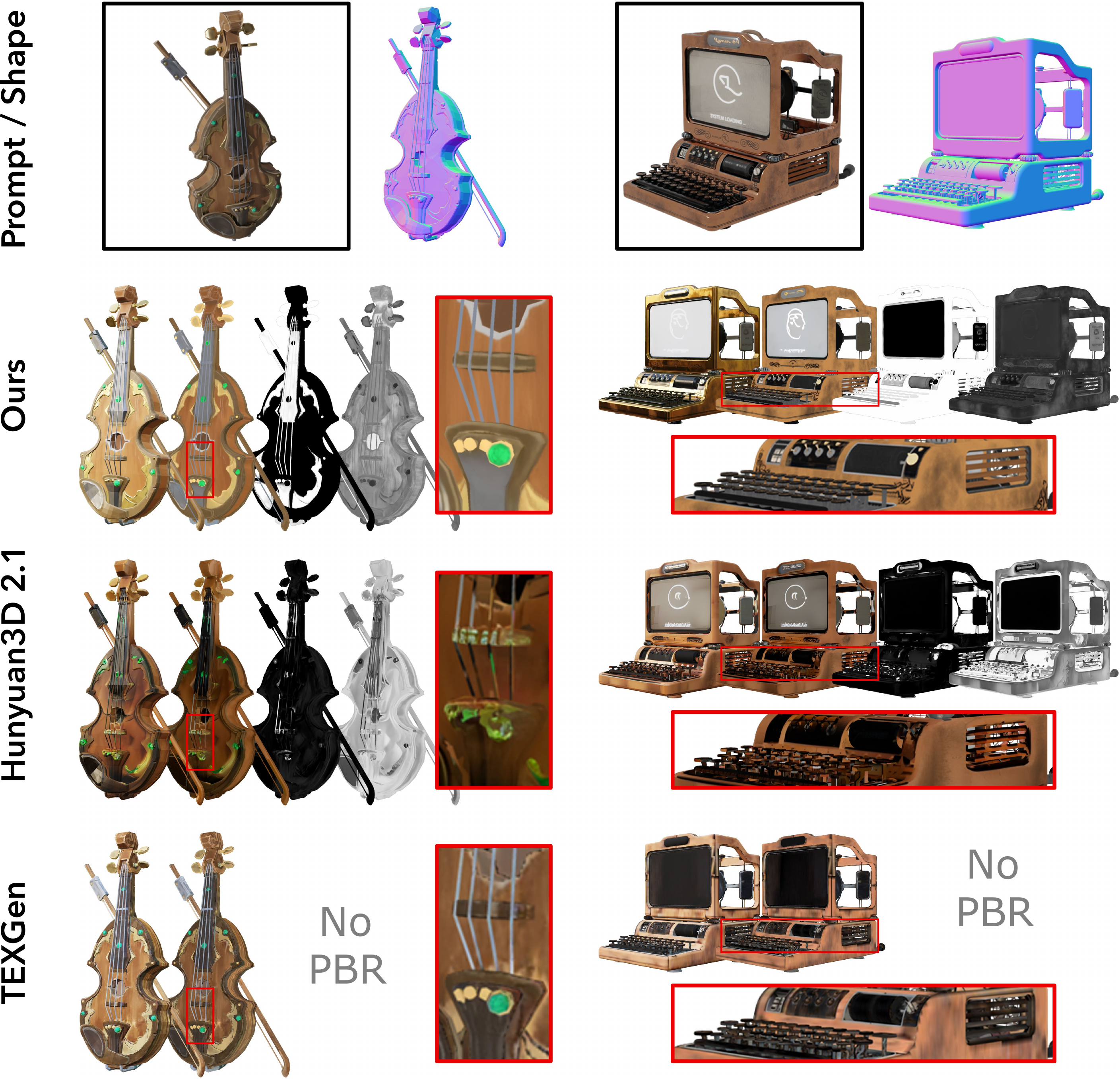}
    \vspace{-20pt}
	\caption{Visual comparison of PBR texture generation.}
	\label{fig:texture}
    \vspace{-8pt}
\end{figure}

\subsection{Shape-Conditioned Texture Generation}

The third stage of our generation pipeline can be independently used as a 3D PBR texture synthesis model given 3D mesh and reference image. To evaluate its performance, we compare against representative baselines: multi-view PBR generation and fusion methods  Hunyuan3D-Paint~\cite{hunyuan3d2025hunyuan3d}, and UV-based TEXGen~\cite{yu2024texgen}.

As shown in Fig.~\ref{fig:texture}, multi-view approaches often suffer from inconsistencies both between the shape and synthesized images, as well as across different views, resulting in ghosting or blurred textures. UV-based methods suffer from ambiguous UV charts and seam artifacts, resulting in degraded visual quality. In contrast, our framework performs appearance reasoning \emph{\textbf{natively in 3D}}. This enables sharper textures, consistent shape–material alignment, and synthesis of textures for internal surfaces, crucial for complex assets with occluded or non-manifold geometry.

\subsection{Ablation and Design Analysis}

We conduct ablation study to analyze the architecture design of SC-VAE. All ablations are conducted on the curated sketchfab assets at a resolution of $256^3$.

\begin{table}[t]  
	\centering  
	\scriptsize
    \caption{Ablation study of SC-VAE architecture designs.}  
    \vspace{-8pt}
	\renewcommand{\arraystretch}{1}
    \setlength{\tabcolsep}{2.5pt}
	\begin{tabular}{lcccccc}  
        \toprule
      \textbf{Setting} & \textbf{\#Token} & \textbf{Dec. {\tiny(ms)}$\downarrow$} & \textbf{MD$\downarrow$} & \textbf{F1\textsubscript{1e-8}$\uparrow$} & \textbf{PSNR$\uparrow$} & \textbf{LPIPS$\downarrow$}  \\  
		 \midrule  
		SC-VAE f16c32 & 503 & \textbf{28.6} & \textbf{1.032} & \textbf{0.312}  & \textbf{27.26} & \textbf{0.072} \\
		  \hspace{5px}\emph{w/o} Residual AE & 503 & 28.7 & 1.747 & 0.268 & 26.73 & 0.081 \\
        \hspace{5px}\emph{w/o} Opt. ResBlock & 503 & 29.6 & 1.198 & 0.285 & 26.67 & 0.083 \\
        \midrule
		SC-VAE f32c128 & 118 & \textbf{33.9} & \textbf{1.405} & \textbf{0.273} & \textbf{26.65} & \textbf{0.081} \\
		  \hspace{5px}\emph{w/o} Residual AE & 118 & 34.0 & 7.394 & 0.192 & 25.01 & 0.102 \\
		\bottomrule
	\end{tabular}  
    \vspace{-8pt}
	\label{tab:ablation}  
\end{table} 

\paravspace
\paragraph{Sparse residual autoencoding.}
To evaluate the effect of the sparse residual autoencoding layer, we compare SC-VAE with a baseline using average pooling and nearest-neighbor upsampling. As shown in Table~\ref{tab:ablation}, the baseline exhibits severe quality degradation: MD increases by 69\% and PSNR decreases 0.5dB at $16\times$ compression, worsening to 526\% and 1.6dB at $32\times$. In contrast, the sparse residual design maintains high fidelity across compression ratios, confirming its robustness under strong spatial bottlenecks.

\paravspace
\paragraph{Optimized residual block.}
To assess the effect of the optimized residual block, we compare SC-VAE with a baseline using standard residual blocks. Table~\ref{tab:ablation} shows that this leads to a clear drop in reconstruction quality, MD increases by 16\% and PSNR decreases by 0.6dB, while runtime remains unchanged. This confirms that the hybrid sparse convolution and point-wise MLP design better preserves fine-scale details without sacrificing efficiency.

\begin{figure}[t]
	\centering
	\includegraphics[width=\linewidth]{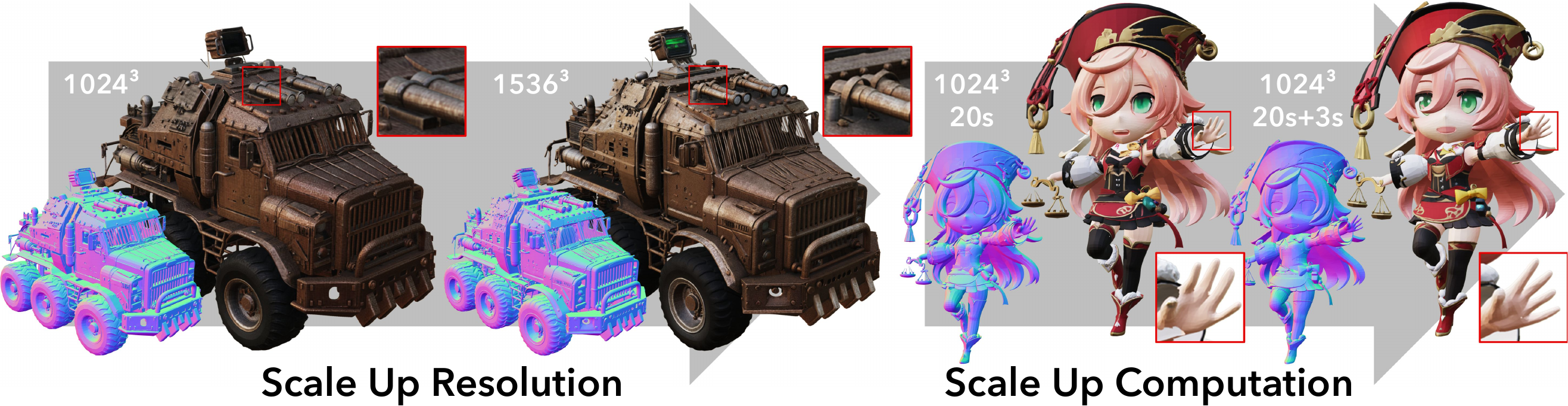}
    \vspace{-16pt}
	\caption{Scaling up resolution for finer detail and compute for higher quality during test time. \textbf{(Best viewed with zoom)}}
	\label{fig:tts}
    \vspace{-4pt}
\end{figure}

\subsection{Test-time Compute and Resolution Scaling}


Our framework enables flexible test-time scaling of both compute and resolution, facilitated by the efficiency of our compact latent space. Notably, our generation process requires significantly fewer tokens than prior approaches, thanks to SC-VAE's $16\times$ spatial compression. This efficiency permits a cascaded application of the second-stage generator, allowing for the synthesis of shapes at resolutions exceeding the training scale efficiently. Specifically, after predicting the O-Voxel structure from the gemetry latent, we can downsample it into a higher-res sparse structure layout (\eg, max-pooling a generated $1024^3$ O-Voxel to a $96^3$ sparse structure resolution). Subsequently, we re-apply the geometry generation stage to obtain a higher-res shape (a $96^3$ sparse structure to a $1536^3$ O-Voxel). See Fig.~\ref{fig:tts} (left) for an exmaple.

Similar stratgies can be applied when operating within the trained resolution to improve generation quality. Rather than directly using the sparse structure generated by the first stage, one can obtain an alternative by downsampling a generated O-Voxel  (\eg, downsampling a $512^3$ O-Voxel to a $64^3$ sparse structure). This can correct local errors and yield cleaner layouts for the subsequent high-resolution generation (\eg, a $1024^3$ O-Voxel). Such a  cascaded inference mechanism offers a controllable trade-off between computational efficiency and generation quality. As demonstrated in Fig.~\ref{fig:tts} (right), cascaded inference yielded finer details and enhanced structural stability.

\section{Conclusion}\label{sec:conclusion}
We present an approach for learning a comprehensive yet compact structured 3D latent representation for 3D generation. A key innovation is O-Voxel, an omni-voxel representation that is capable of encoding complex geometry and materials. We also introduce a Sparse Compression VAE that achieves a high spatial compression rate on O-Voxel to construct the latent space. Our large flow-matching models deliver substantially superior generation quality  compared to existing methods while maintaining high efficiency. We believe our approach offers a significant advancement in 3D generative modeling, enhancing both the efficiency and realism of 3D content creation and opening avenues for broader applications across various domains.

{\small
\bibliographystyle{ieeenat_fullname}
\bibliography{ref}
}

\clearpage
\appendix
\pagenumbering{arabic}
\setcounter{page}{1}

\section{More Implementation Details}

\subsection{O-Voxel Conversion Algorithms}

This section provides a detailed breakdown of the bidirectional conversion algorithms between standard 3D assets (meshes and PBR textures) and our O-Voxel representation. We present the process in four parts: converting a mesh to the O-Voxel shape representation, reconstructing a mesh from it, converting PBR textures to the O-Voxel material representation, and reconstructing textures from it.

\subsubsection{Shape Conversion}

The geometric component of O-Voxel is based on our \emph{Flexible Dual Grid} formulation. Algo.~\ref{alg:mesh_to_ovoxel_shape} and \ref{alg:ovoxel_to_mesh_shape} detail its conversion to and from a triangle mesh.

\subsubsection{Material Conversion}
The material component of O-Voxel stores PBR attributes in active voxels. The conversion to and from standard mesh textures is a direct sampling and interpolation process (Algo.~\ref{alg:texture_to_ovoxel_material} and \ref{alg:ovoxel_to_texture_material}).

\begin{algorithm}
    \caption{Mesh-to-O-Voxel Conversion}
    \label{alg:mesh_to_ovoxel_shape}
    \DontPrintSemicolon
    \KwInput{Input mesh $\mathcal{M}$, grid resolution $N$, weights $\lambda_{\text{bound}}, \lambda_{\text{reg}}$}
    \KwOutput{O-Voxel shape features $\boldsymbol{f}^{\text{shape}}$}
    
    \BlankLine
    \Comment{1. Initialize a map to store data for each active voxel}
    $\text{voxel\_data} \leftarrow \text{EmptyMap}[\boldsymbol{p} \rightarrow \{\text{QEF}, \bar{\boldsymbol{q}}, \boldsymbol{\delta}\}]$
    
    \BlankLine
    \Comment{2. Accumulate plane-distance QEFs from triangle intersections}
    \For{each triangle $T$ in $\mathcal{M}$}{
        \For{each voxel edge $e$ intersected by $T$}{
            $\{\boldsymbol{q}, \boldsymbol{n}\} \leftarrow$ GetIntersectionAndNormal($T, e$)
            
            \For{each neighboring voxel $V$ of edge $e$}{
                $\boldsymbol{p} \leftarrow$ GetCoordinate($V$)\;
                Initialize $\text{voxel\_data}[\boldsymbol{p}]$ if not exists\;
                $\text{plane\_qef} \leftarrow$ BuildPlaneQEF($\boldsymbol{q}, \boldsymbol{n}$)\;
                $\text{voxel\_data}[\boldsymbol{p}].\text{QEF}.\text{acc}(\text{plane\_qef})$\;
                $\text{voxel\_data}[\boldsymbol{p}].\bar{\boldsymbol{q}}.\text{acc}(\boldsymbol{q})$\;
                $\text{voxel\_data}[\boldsymbol{p}].\boldsymbol{\delta}.\text{update}(e)$
            }
        }
    }

    \BlankLine
    \Comment{3. Accumulate boundary-distance QEFs from open mesh edges}
    \For{each boundary edge $b$ in $\mathcal{M}$}{
        \For{each voxel $V$ intersected by $b$}{
            $\boldsymbol{p} \leftarrow$ GetCoordinate($V$)\;
            \If{$\boldsymbol{p}$ in $\text{voxel\_data}$}{
                $\{\boldsymbol{o}, \boldsymbol{d}\} \leftarrow$ GetLineParameters($b$)\;
                $\text{line\_qef} \leftarrow$ BuildLineQEF($\boldsymbol{o}, \boldsymbol{d}$)\;
                $\text{voxel\_data}[\boldsymbol{p}].\text{QEF}.\text{acc}(\lambda_{\text{bound}} \cdot \text{line\_qef})$
            }
        }
    }

    \BlankLine
    \Comment{4. Accumulate regularization-term QEFs}
    \For{each $\boldsymbol{p}$ in $\text{voxel\_data}$}{
        $\text{reg\_qef} \leftarrow$ BuildPointQEF($\text{voxel\_data}[\boldsymbol{p}].\bar{\boldsymbol{q}}$)\;
        $\text{voxel\_data}[\boldsymbol{p}].\text{QEF}.\text{acc}(\lambda_{\text{reg}} \cdot \text{reg\_qef})$
    }
    
    \BlankLine
    \Comment{5. Solve QEFs and finalize O-Voxel features}
    $\boldsymbol{f}^{\text{shape}} \leftarrow \text{EmptyMap}[]$\;
    \For{each $\boldsymbol{p}$, data in $\text{voxel\_data}$}{
        $\boldsymbol{v} \leftarrow$ SolveQEF($\text{data.QEF}$) \;
        $\boldsymbol{\delta} \leftarrow \text{data}.\boldsymbol{\delta}$ \;
        $\gamma \leftarrow 0.5$\;
        
        $\boldsymbol{f}^{\text{shape}}[\boldsymbol{p}] \leftarrow \{\boldsymbol{v}, \boldsymbol{\delta}, \gamma\}$
    }
    \Return $\boldsymbol{f}^{\text{shape}}$
\end{algorithm}

\begin{algorithm}
    \caption{O-Voxel-to-Mesh Conversion}
    \label{alg:ovoxel_to_mesh_shape}
    \DontPrintSemicolon
    \KwInput{O-Voxel shape features $\boldsymbol{f}^{\text{shape}}$}
    \KwOutput{Reconstructed mesh $\mathcal{M}'$}
    
    \BlankLine
    \Comment{1. Create a mesh vertex for each dual vertex in the O-Voxel data}
    $V' \leftarrow \text{EmptyList}[]$\;
    $\text{vertex\_indices} \leftarrow \text{EmptyMap}[\boldsymbol{p} \rightarrow \text{index}]$\;
    \For{each $\boldsymbol{p}$, data in $\boldsymbol{f}^{\text{shape}}$}{
        $V'.\text{append}(\text{data}.\boldsymbol{v})$\;
        $\text{vertex\_indices}[\boldsymbol{p}] \leftarrow |V'| - 1$\;
    }

    \BlankLine
    \Comment{2. Generate faces by connecting vertices across active edges}
    $F' \leftarrow \text{EmptyList}[]$\;
    \For{each $\boldsymbol{p}$, data in $\boldsymbol{f}^{\text{shape}}$}{
        \Comment{Iterate over the 3 predefined axes to avoid duplicate faces}
        \For{each axis $a \in \{X, Y, Z\}$}{
            \If{$\text{data}.\boldsymbol{\delta}[a] == 1$}{
                $\text{quad\_coords} \leftarrow$ GetQuadVoxel($\boldsymbol{p}$, $a$)\;
                
                \Comment{Ensure all four voxels are active}
                \If{all $\text{quad\_coords}$ exist in $\boldsymbol{f}^{\text{shape}}$}{
                    $i_0, i_1, i_2, i_3 \leftarrow \text{vertex\_indices}[\text{quad\_coords}]$\;
                    \Comment{Split the quadrilateral into two triangles}
                    $t_1, t_2 \leftarrow$ Split($\{i_0, i_1, i_2, i_3\}$, $\text{data}.\gamma$)\;
                    $F'.\text{extend}([t_1, t_2])$\;
                }
            }
        }
    }
    
    \BlankLine
    \Comment{3. Construct the final mesh from vertices and faces}
    $\mathcal{M}' \leftarrow \text{Mesh}(V', F')$\;
    \Return $\mathcal{M}'$
\end{algorithm}

\begin{algorithm}
    \caption{Texture-to-O-Voxel Conversion}
    \label{alg:texture_to_ovoxel_material}
    \DontPrintSemicolon
    \KwInput{Input mesh $\mathcal{M}$ with PBR textures, O-Voxel shape features $\boldsymbol{f}^{\text{shape}}$}
    \KwOutput{O-Voxel material features $\boldsymbol{f}^{\text{mat}}$}
    
    \BlankLine
    \Comment{1. Initialize an empty map for material features}
    $\boldsymbol{f}^{\text{mat}} \leftarrow \text{EmptyMap}[\boldsymbol{p} \rightarrow \{\boldsymbol{c},m,r,\alpha\}]$
    
    \BlankLine
    \Comment{2. For each active voxel, sample material attributes from the mesh}
    \For{each voxel coordinate $\boldsymbol{p}$ in $\boldsymbol{f}^{\text{shape}}$}{
        $\boldsymbol{p}_{\text{center}} \leftarrow$ GetVoxelCenter($\boldsymbol{p}$) \;
        $\text{intersecting\_tris} \leftarrow$ FindIntersectingTriangles($\mathcal{M}$, $\boldsymbol{p}$) \;
        \BlankLine
        \Comment{Collect weighted samples from all intersecting triangles}
        $\text{samples} \leftarrow \text{EmptyList}[]$ \;
        $\text{weights} \leftarrow \text{EmptyList}[]$ \;
        \For{each triangle $T$ in $\text{intersecting\_tris}$}{
            $\boldsymbol{q} \leftarrow$ ProjectPointOntoTriangle($\boldsymbol{p}_{\text{center}}$, $T$) \;
            $d \leftarrow \|\boldsymbol{p}_{\text{center}}, \boldsymbol{q}\|_2$ \;
            $w \leftarrow 1 - d$ \;
            $\text{mip\_level} \leftarrow$ GetMipLevel($T$, $\text{voxel\_size}$) \;
            $\boldsymbol{uv} \leftarrow$ GetUVCoordinates($\boldsymbol{q}$, $T$) \;
            $\text{pbr\_sample} \leftarrow$ SampleTexture($\mathcal{M}.\text{textures}$, $\boldsymbol{uv}$, $\text{mip\_level}$) \;
            
            $\text{samples.append}(\text{pbr\_sample})$ \;
            $\text{weights.append}(w)$ \;
        }
        
        \BlankLine
        \Comment{Compute the final feature via weighted average}
        $\boldsymbol{f}^{\text{mat}}[\boldsymbol{p}] \leftarrow $WeightedAverage($\text{samples}$, $\text{weights}$) \;
    }
    \BlankLine
    \Return $\boldsymbol{f}^{\text{mat}}$
\end{algorithm}

\begin{algorithm}
    \caption{O-Voxel-to-Texture Conversion}
    \label{alg:ovoxel_to_texture_material}
    \DontPrintSemicolon
    \KwInput{Reconstructed mesh $\mathcal{M}'$, O-Voxel material features $\boldsymbol{f}^{\text{mat}}$, mode $\in \{\text{'vertex', 'map'}\}$}
    \KwOutput{Mesh $\mathcal{M}'$ with PBR materials applied}
    
    \BlankLine
    \If{mode == 'vertex'}{
        \Comment{1. Generate vertex colors via trilinear interpolation}
        $\text{vertex\_materials} \leftarrow \text{EmptyList}[]$\;
        \For{each vertex $\boldsymbol{v}$ in $\mathcal{M}'$}{
            $\text{material} \leftarrow$ TrilinearInterp($\boldsymbol{v}$, $\boldsymbol{f}^{\text{mat}}$)\;
            $\text{vertex\_materials.append}(\text{material})$\;
        }
        ApplyVertexMaterials($\mathcal{M}'$, $\text{vertex\_materials}$)\;
    }
    \ElseIf{mode == 'map'}{
        \Comment{2. Generate texture maps by filling interpolated values}
        $\text{texture\_maps} \leftarrow$ Parameterize($\mathcal{M}'$)\;
        
        \For{each texel $(u,v)$ in $\text{texture\_maps}$}{
            $\boldsymbol{q} \leftarrow$ GetSurfacePointFromUV($\mathcal{M}'$, $u, v$)\;
            $\text{material} \leftarrow$ TrilinearInterp($\boldsymbol{q}$, $\boldsymbol{f}^{\text{mat}}$)\;
            $\text{texture\_maps}[u,v] \leftarrow \text{material}$\;
        }
        ApplyTextureMaps($\mathcal{M}'$, $\text{texture\_maps}$)\;
    }
    
    \BlankLine
    \Return $\mathcal{M}'$
\end{algorithm}

\subsection{Network Architectures}

\begin{table}
    \centering  
    \scriptsize
    \caption{Architectural details of the SC-VAE encoder. The decoder follows a symmetrical design.}  
    \vspace{-8pt}
	\begin{tabular}{c|c}  
        \toprule
        \textbf{Stage ($\boldsymbol{f_{\text{down}}}$)} & \textbf{Block} \\
        \midrule  
            \multirow{2}{*}{\textbf{1$\times$}} & $\operatorname{Linear}(6, 64)$ \\
            & $\operatorname{ResEnc}(64, 128)$ \\
        \midrule  
            \multirow{2}{*}{\textbf{2$\times$}} & $\left[\begin{matrix}\operatorname{SubMConv(3, 128, 128)}\\\operatorname{LayerNorm}\\\operatorname{Linear}(128,512)\\\operatorname{SiLU}\\\operatorname{Linear}(512,128)\end{matrix}\right]\times4$  \\
            & $\operatorname{ResEnc}(128, 256)$ \\
        \midrule  
            \multirow{2}{*}{\textbf{4$\times$}} & $\left[\begin{matrix}\operatorname{SubMConv(3, 256, 256)}\\\operatorname{LayerNorm}\\\operatorname{Linear}(256,1024)\\\operatorname{SiLU}\\\operatorname{Linear}(1024,256)\end{matrix}\right]\times8$  \\
            & $\operatorname{ResEnc}(256, 512)$ \\
        \midrule  
            \multirow{2}{*}{\textbf{8$\times$}} & $\left[\begin{matrix}\operatorname{SubMConv(3, 512, 512)}\\\operatorname{LayerNorm}\\\operatorname{Linear}(512,2048)\\\operatorname{SiLU}\\\operatorname{Linear}(2048,512)\end{matrix}\right]\times16$  \\
            & $\operatorname{ResEnc}(512, 1024)$ \\
        \midrule  
            \multirow{2}{*}{\textbf{16$\times$}} & $\left[\begin{matrix}\operatorname{SubMConv(3, 1024, 1024)}\\\operatorname{LayerNorm}\\\operatorname{Linear}(1024,4096)\\\operatorname{SiLU}\\\operatorname{Linear}(4096,1024)\end{matrix}\right]\times4$  \\
            & $\operatorname{Linear}(1024, 32\times2)$ \\
        \bottomrule
    \end{tabular}  
    \label{tab:scvae_detail} 
\end{table}

\paragraph{Sparse Compression VAE} The Sparse Compression VAE (SC-VAE) is a \emph{fully sparse-convolutional network} designed to compress the O-Voxel representation into a compact latent space with a 16$\times$ spatial downsampling ratio. We employ a conventional U-Shaped VAE architecture, optimized with ConvNeXt-style~\cite{liu2022convnet} residual blocks and Residual AutoEncoding layers~\cite{chen2024deep} for (down/up)sampling. The detailed architecture for the SC-VAE encoder is presented in Table\ref{tab:scvae_detail}. The decoder is constructed symmetrically using inverted block numbers and dimensions.
The complete model comprises $\sim$800M parameters (354M for the encoder and 474M for the decoder). This configuration achieves near-lossless reconstruction fidelity while maintaining high computational efficiency.

\paravspace
\paragraph{Generative Models.}
Our generation framework consists of three Transformer-based models. These models adopt a standard encoder-only architecture, intentionally omitting complex designs such as token packing or skip connections to maintain a clean and scalable architecture (shown in Table~\ref{tab:dit_detail}). Conditional inputs are integrated using  mechanisms tailored to the nature of the data:
\begin{itemize}
    \item \textbf{Timestep Injection:} We use the \emph{AdaLN-single}~\cite{chen2024pixart} scheme for the conditioning on diffusion timestep. This method modulates the activations within the network, allowing the model to effectively incorporate temporal information while drastically reducing required parameters comparing to the AdaLN baseline.
    \item \textbf{Image Prompt Conditioning:} Image prompts are integrated via \emph{cross-attention} layers. This enables the model to align its generative process with the semantic content of the visual conditioning signal.
    \item \textbf{Shape Conditioning:} For the material generation stage, shape information is provided as a condition by \emph{concatenating} it channel-wise with the input tensor. This direct approach ensures that geometric constraints are explicitly provided, help improve material-shape alignment.
\end{itemize}

\noindent To enhance generalization across different input resolutions, we incorporate \emph{Rotary Position Embedding (RoPE)}~\cite{su2024roformer}. Furthermore, we employ a \emph{QK-Norm} scheme~\cite{esser2024scaling} to stabilize the attention mechanism. This involves applying Root Mean Square Normalization (RMSNorm)~\cite{zhang2019root} to the query and key tensors before the attention operation, which improves training stability.

\begin{table}
    \centering  
    \scriptsize
    \caption{Architectural Details for the Generative models.}  
    \vspace{-8pt}
	\begin{tabular}{c|c}  
        \toprule
        \textbf{Stage} & \textbf{Block} \\
        \midrule  
            \textbf{In\_proj} & $\operatorname{Linear}(32(+32), 1536)$ \\
        \midrule  
            \textbf{Stem} & $\left[\begin{matrix}\operatorname{AdaLN-single}\\\operatorname{SelfAttn}( 12\times128)\\\operatorname{LayerNorm}\\\operatorname{CrossAttn}(12\times128)\\\operatorname{AdaLN-single}\\\operatorname{FFN}(1536,8192)\end{matrix}\right]\times30$  \\
        \midrule  
            \multirow{2}{*}{\textbf{Out\_proj}} & $\operatorname{LayerNorm}$ \\
            & $\operatorname{Linear}(1536, 32)$ \\
        \bottomrule
    \end{tabular}  
    \label{tab:dit_detail} 
\end{table}

\subsection{Training Details}

\paragraph{Sparse Compression VAE.}
As described in the main paper, the SC-VAE is trained using a two-stage strategy. The first stage focuses on stabilizing the training process by employing a direct O-Voxel feature regression loss at a resolution of $256^3$. In the second stage, resolution is increased to $512^3$ while rendering-based perceptual loss is introduced to enhance visual quality, such as geometric sharpness and high-frequency material details, and to facilitate the model's adaptation to higher resolutions. This rendering loss is implemented as follows:
\begin{equation}
	\begin{split}
        d_{\text{p}}(\boldsymbol{a},\boldsymbol{b}) &= \|\boldsymbol{a}-\boldsymbol{b}\|_1 + 0.2 \cdot d_{\text{SSIM}} + 0.2 \cdot d_{\text{LPIPS}} \\
		\mathcal{L}^{\text{shape}}_{\text{render}} &= \|\hat{m}-m\|_1 + 10 \cdot \|\hat{d}-d\|_1 + d_{\text{p}}(\hat{\boldsymbol{n}}, \boldsymbol{n}) \\
        \mathcal{L}^{\text{mat}}_{\text{render}} &= d_{\text{p}}(\hat{\boldsymbol{c}},\boldsymbol{c}) + d_{\text{p}}(\hat{\boldsymbol{mra}},\boldsymbol{mra})
	\end{split}
	\label{eq:rendering_loss}
\end{equation}
where $\mathcal{L}^{\text{shape}}_{\text{render}}$ and $\mathcal{L}^{\text{mat}}_{\text{render}}$ are the rendering losses for shape and material, respectively. The term $d_{\text{p}}(\cdot, \cdot)$ denotes a perceptual distance metric combining the L1 norm with SSIM and LPIPS losses. In these equations, variables with a hat ($\hat{\cdot}$) represent model predictions, while variables without are the ground-truth targets. Specifically, $m$ is the silhouette mask, $d$ is the depth map, $\boldsymbol{n}$ is the normal map, $\boldsymbol{c}$ is the base color, and $\boldsymbol{mra}$ corresponds to the metallic-roughness-alpha map.

\emph{For inputs with resolutions exceeding $512^3$, we directly apply the pre-trained SC-VAE models without modification.} The fully sparse-convolutional design of the SC-VAE is inherently resolution-agnostic, a property that allows the models to generalize effectively to larger spatial resolutions without requiring fine-tuning.

\paravspace
\paragraph{Generative Models.}
We employ the \emph{rectified flow} formulation~\cite{liu2023flow} to train our generative models. This framework defines a forward process based on linear interpolation, $\boldsymbol{x}(t)=(1-t)\boldsymbol{x}_0+t\boldsymbol{\epsilon}$, which constructs a straight path from a data sample $\boldsymbol{x}_0$ to a random noise sample $\boldsymbol{\epsilon}$, indexed by timestep $t \in [0, 1]$.

The corresponding reverse process is governed by a time-dependent vector field, $\boldsymbol{v}(\boldsymbol{x}, t) = \nabla_t\boldsymbol{x}$, which guides samples from the noise distribution back toward the data distribution. This vector field is approximated by a neural network, denoted $\boldsymbol{v}_\theta$, which is trained by minimizing the \emph{Conditional Flow Matching (CFM)} objective~\cite{lipman2023flow}:
\begin{equation}
    \mathcal{L}_{\text{CFM}}(\theta)=\mathbb{E}_{t,\boldsymbol{x}_0,\boldsymbol{\epsilon}}\|\boldsymbol{v}_\theta(\boldsymbol{x}(t), t)-(\boldsymbol{\epsilon}-\boldsymbol{x}_0)\|^2_2.
    \label{eq:cfm}
\end{equation}
Following the approach of~\cite{xiang2025structured}, we adopt an altered timestep sampling strategy, utilizing a $\mathrm{logitNorm}(1,1)$ distribution for better generation quality.

\section{FlexGEMM: Our High-Performance Sparse Convolution Backend}

\begin{figure*}[t]
	\centering
	\includegraphics[width=\linewidth]{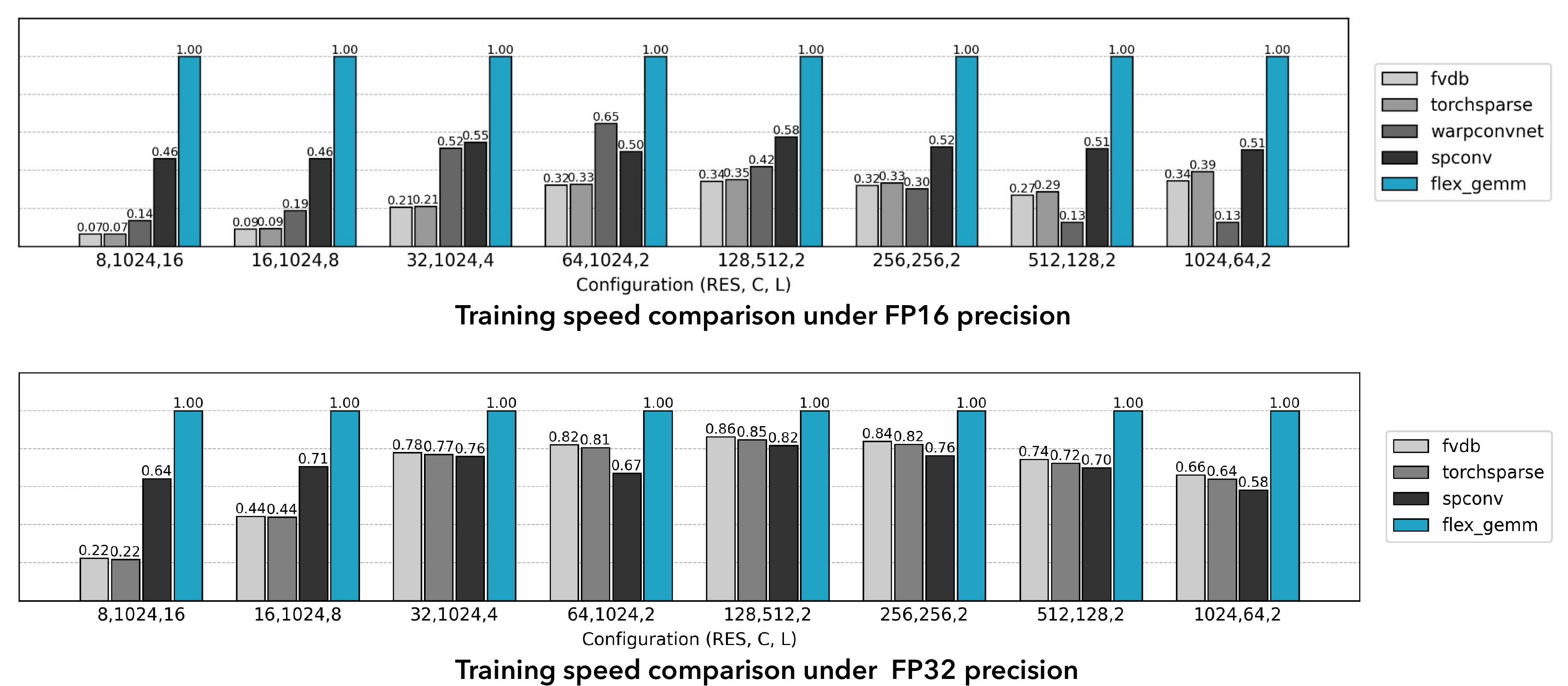}
    \vspace{-16pt}
	\caption{Speed test for FlexGEMM backend and baselines including Spconv~\cite{spconv2022}, Torchsparse~\cite{tangandyang2023torchsparse}, fvdb~\cite{williams2024fvdb}, and WarpConvNet~\cite{warpconvnet2025}}
	\label{fig:spconv_bench}
    \vspace{-8pt}
\end{figure*}

The sparse convolutional networks in our model are accelerated by a \emph{custom} high-performance backend developed for this work. This backend was engineered to address the performance and platform-dependency limitations of existing libraries, which are often tightly coupled to the NVIDIA CUDA ecosystem. By implementing our kernels in \emph{Triton}~\cite{tillet2019triton}, a high-level GPU programming language, we created a single, cross-platform codebase that delivers near-optimal performance on both NVIDIA and AMD hardware.

Our final, optimized implementation employs a \emph{Masked Implicit GEMM} strategy~\cite{spconv2022}. This approach moves beyond naive explicit matrix multiplication by fusing the feature gathering (im2col) and the matrix multiplication (GEMM) steps into a single, highly-optimized kernel. This fusion minimizes global memory I/O by keeping intermediate data in fast on-chip memory. To further enhance performance in sparse contexts, we introduce a masking mechanism that intelligently skips computation on empty neighbor slots. This is achieved by first reordering active voxels using \emph{Gray code ordering}, a technique that groups voxels with similar neighborhood patterns together. This grouping significantly improves the SIMD efficiency of the GPU, reducing warp divergence and wasted computation. Finally, we incorporate a \emph{Split-K} technique, which increases parallelism by dividing the accumulation dimension of the matrix multiplication into independent parallel tasks. This is particularly effective in common scenarios, such as those with a large number of channels or a small number of active voxels. The combination of these techniques results in a highly efficient backend, yielding up to a \emph{2$\times$ speedup over widely-used sparse convolution libraries} in our benchmarking (see Fig.~\ref{fig:spconv_bench}).

\section{Data Preparation Details}

\begin{table}[t]
    \centering  
    \scriptsize
    \caption{Composition of the training set and evaluation set.}  
    \vspace{-8pt}
	\begin{tabular}{c|cc}  
        \toprule
        \textbf{Source} & \textbf{Shape Availible} & \textbf{Material Availible} \\
        \midrule  
        TexVerse~\cite{zhang2025texverse} & 503387 & 382996 \\
        ObjaverseXL (sketchfab)~\cite{deitke2024objaverse} & 168307 & 141623 \\
        ObjaverseXL (github)~\cite{deitke2024objaverse} & 293887 & 202188 \\
        ABO~\cite{collins2022abo} & 4485 & 4485 \\
        HSSD~\cite{khanna2023hssd} & 6670 & 6670 \\
        \textbf{SC-VAE training set} & 473349 & 354966 \\
        \textbf{All training set} & 976736 & 737962 \\
        \midrule  
        Toys4k~\cite{stojanov2021using} (evaluation set) & 3229 & 2282 \\
        \bottomrule
    \end{tabular}  
    \label{tab:dataset} 
\end{table}

The data preparation pipeline is largely based on the setup proposed in \textsc{Trellis}~\cite{xiang2025structured}. We begin by curating a collection of 3D assets but exclude the 3D-FUTURE~\cite{fu20213d} dataset due to its lack of Physically-Based Rendering (PBR) materials. The remaining assets form the basis for training our SC-VAEs.

All assets in this curated collection are used to extract geometric data for training the shape SC-VAE. For the material SC-VAE, a more specific filtering process is required. We employ a custom Blender~\cite{blender} script to parse materials from the raw assets and retain only those that utilize a standard metallic-roughness PBR workflow. This filtering process yields a subset of approximately 350,000 assets suitable for training the material VAE.

To train the generative models, we further augment the dataset with TexVerse~\cite{zhang2025texverse} to increase the diversity of high-quality PBR materials. As a final quality control step, we filter the assets based on an aesthetic score. For simplicity, we leverage the thumbnail images provided on the Sketchfab~\cite{sketchfab2025} platform to estimate this score. Objects with an estimated aesthetic score below 4.5 are excluded from the training set. Detailed statistics of the final dataset are provided in Table~\ref{tab:dataset}.

To generate the image prompts required for training our image-conditioned model, we render a diverse set of views for each 3D asset using Blender. We apply a series of augmentations during this rendering process to ensure the model is robust against common ambiguities found in real-world inputs. Key augmentations include:
\begin{itemize}
    \item \textbf{Field of View (FoV):} The camera's Field of View (FoV) is randomly sampled between $10^\circ$ and $70^\circ$. This augmentation is designed to make the model robust to variations in camera intrinsics, which are often unknown in practice.
    
    \item \textbf{Lighting Conditions:} The lighting environment is randomized by ramdomly placing and adjusting the intensity of light sources. This improves the model's ability to predict intrinsic PBR attributes accurately, disentangling them from environmental illumination.
\end{itemize}

\section{More Experiment Details}

\subsection{Evaluation Protocol}
In the main paper, we present quantitative comparisons and ablation studies using a series of numerical metrics. We provide the detailed protocols for their calculation below.

\subsubsection{Reconstruction Experiments}

\paragraph{Test set.}
To ensure a robust evaluation of reconstruction quality, we prepared two distinct test sets.
\begin{itemize}
    \item \textit{Toys4k-PBR.} Our first test set is derived from the Toys4k dataset. For a rigorous metric, we filtered the raw assets to include only those containing all three standard PBR maps (base color, metallic, and roughness). This process resulted in a refined test set of 473 instances.
    \item \textit{Sketchfab Featured.} Recognizing that the assets in Toys4k are relatively simple, we curated a second, more challenging test set from high-quality, recent assets on Sketchfab. Specifically, we selected models from the ``Staff Picks" category, which features professionally curated content. We then applied a filter to retain only assets that utilize the metallic-roughness PBR workflow and were uploaded within the last two years. This process yielded a high-quality test set comprising 90 instances, designed to evaluate performance on complex, professional-grade assets.
\end{itemize}

\paravspace
\paragraph{Geometry Accuracy.}
To assess the overall geometric fidelity, we use \emph{Mesh Distance} and the corresponding \emph{F-score}. Unlike Chamfer Distance, which is sensitive to point cloud density, Mesh Distance provides a more stable measure of the discrepancy between two triangle meshes. This makes it particularly suitable for evaluating reconstruction accuracy across all surfaces, including those that are fully enclosed. For this evaluation, we sample 1 million points from the surface of each mesh. For the F-score calculation, we use a distance threshold of $\tau = 1 \times 10^{-8}$.

For evaluating the accuracy of visible surfaces, we compute \emph{Chamfer Distance (CD)} and the corresponding \emph{F-score}. The evaluation is performed on point clouds generated by sampling the outer shell of the meshes. Specifically, we render depth maps for each mesh from 100 uniformly sampled camera views. These depth maps are then unprojected to create a dense 3D point cloud, from which we randomly sample 1 million points. For the F-score calculation, we use a distance threshold of $\tau = 1 \times 10^{-6}$.

To evaluate the quality of fine surface details, we compute \emph{PSNR} and \emph{LPIPS} on rendered normal maps. For this, we render images from four fixed camera positions for all assets. The camera is placed on a sphere of radius 10 with a fixed pitch angle of $30^\circ$ and a narrow Field of View (FoV) of $6^\circ$. The four views correspond to yaw angles of $30^\circ, 120^\circ, 210^\circ,$ and $300^\circ$.

Prior to any metric calculation, all ground-truth and predicted meshes are normalized to fit within a unit cube. The definitions for the geometric metrics are as follows:
\begin{itemize}[leftmargin=2em]
    \item \emph{Mesh Distance (MD).} MD is calculated as the bidirectional point-to-mesh surface distance, averaged over a dense sampling of points from both meshes. Given two meshes $S_X$ and $S_Y$, with sampled points $P_X$ and $P_Y$, MD is defined as:
    \begin{equation}
    \begin{split}
    \text{MD}(S_X, S_Y) = &\frac{1}{2|P_X|} \sum_{\boldsymbol{x} \in P_X} \min_{\boldsymbol{y} \in S_Y} \|\boldsymbol{x} - \boldsymbol{y}\|_2^2 
    \\ + &\frac{1}{2|P_Y|} \sum_{\boldsymbol{y} \in P_Y} \min_{\boldsymbol{x} \in S_X} \|\boldsymbol{y} - \boldsymbol{x}\|^2_2.
    \end{split}
    \label{eq:mesh_distance}
    \end{equation}

    \item \emph{Chamfer Distance (CD).} Given two point clouds, $X$ and $Y$, the Chamfer Distance is defined as:
    \begin{equation}
    \begin{split}
        \text{CD}(X,Y) = &\frac{1}{2|X|}\sum_{\boldsymbol{x}\in{X}}\min_{\boldsymbol{y}\in{Y}}\|\boldsymbol{x}-\boldsymbol{y}\|_2^2
        \\+ &\frac{1}{2|{Y}|}\sum_{\boldsymbol{y}\in{Y}}\min_{\boldsymbol{x}\in{X}}\|\boldsymbol{y}-\boldsymbol{x}\|_2^2.
    \end{split}
    \end{equation}

    \item \emph{F-score.} The F-score evaluates shape correspondence by combining precision and recall, calculated based on a distance threshold $\tau$. Given a ground-truth shape $S_{gt}$ and a predicted shape $S_{pred}$, we sample point sets $P_{gt}$ from $S_{gt}$ and $P_{pred}$ from $S_{pred}$. Precision and Recall are then defined as:
    \begin{equation}
    \begin{split}
        \text{Prec}(\tau) &= \frac{1}{|P_{pred}|} \sum_{p \in P_{pred}} \mathbb{I}\left( d^2(p, S_{gt}) < \tau \right), \\
        \text{Rec}(\tau) &= \frac{1}{|P_{gt}|} \sum_{p \in P_{gt}} \mathbb{I}\left( d^2(p, S_{pred}) < \tau \right),
    \end{split}
    \end{equation}
    where $\mathbb{I}(\cdot)$ is the indicator function and $d(p, S)$ is the minimum Euclidean distance from a point $p$ to the shape $S$. The F-score is the harmonic mean of these values:
    \begin{equation}
        \mathbf{F\text{-score}}(\tau) = \frac{2 \cdot \text{Prec}(\tau) \cdot \text{Rec}(\tau)}{\text{Prec}(\tau) + \text{Rec}(\tau)}.
    \end{equation}
    The distance function $d(p, S)$ is defined differently depending on the context, described below.
    \begin{itemize}
        \item For CD F-score: The shapes $S_{gt}$ and $S_{pred}$ are treated as discrete point clouds. The distance $d(p, S)$ is the Euclidean distance from point $p$ to the nearest point within the point cloud $S$.
        \item For MD F-score: The shapes $S_{gt}$ and $S_{pred}$ are treated as continuous triangle meshes. The distance $d(p, S)$ is the Euclidean distance from point $p$ to the closest point on the surface of the mesh $S$.
    \end{itemize}
\end{itemize}

\paravspace
\paragraph{Appearance Fidelity.}
To assess the quality of the reconstructed materials, we evaluate both the raw PBR attribute maps and the final shaded images. For both the ground-truth and the reconstructed 3D assets, we render two sets of images using the nvdiffrec renderer~\cite{munkberg2022extracting}. This rendering is performed using the same fixed-camera setup as the normal map evaluation, capturing four distinct views. The PSNR and LPIPS metrics are then calculated by comparing the rendered outputs from the reconstructed asset against those from the ground truth. The final reported scores are the average values across these four views.

\subsubsection{Generation Experiments}

\paragraph{Test Set.}
For quantitative evaluation of our image-to-3D generation capabilities, we conduct experiments on a challenging test set of 100 image prompts generated by the NanoBanana text-to-image model~\cite{fortin2025nanobanana}. This dataset was specifically chosen for its diversity and complexity. It features prompts that describe objects with intricate geometries, varied and dramatic lighting conditions, and a wide range of realistic materials, including metal, leather, rust, and translucent substances such as glass.

\paragraph{Evaluation Metrics.}
We employ a suite of metrics targeting different aspects of the output. The visual and semantic alignment between the input image prompt and rendered images of the asset is measured using the \emph{CLIP score}~\cite{radford2021learning}. To evaluate how well the 3D geometry and appearance properties match the image prompt, we use the multimodal foundation models \emph{ULIP-2}~\cite{xue2024ulip} and \emph{Uni3D}~\cite{zhou2023uni3d}. Details of the metrics are listed below:
\begin{itemize}
    \item \emph{CLIP Score.} The CLIP score measures the semantic similarity between two images. In our evaluation, we render the generated 3D asset from 4 predefined viewpoints (same yaw, pitch setup as previous metrics). We then compute the average cosine similarity between the CLIP embedding of the input image prompt and the rendered images (or normal map). A higher CLIP score indicates a better semantic alignment between the conditional input and the appearance (or geometry) of the generated asset.
    \item \emph{ULIP-2 and Uni3D Scores.} ULIP-2 and Uni3D are models designed to understand and align 3D content with text/image . To prepare the input for these models, we first convert our generated mesh into a colored point cloud. Specifically, we uniformly sample 10,000 points from the surface of the mesh with Farthest Point Sampling. The color for each point is determined by querying its corresponding RGB value from the asset's base color map. This colored point cloud is then fed into the ULIP-2 and Uni3D models to compute a similarity score against the image prompt. These scores provide a quantitative measure of how well the generated asset align with the condition from a native 3D perspective.
\end{itemize}

\subsection{User Study}

While quantitative metrics provide objective measurements of fidelity, they often fail to capture the nuanced perceptual qualities that define a high-quality 3D asset, such as aesthetic appeal and fine-detail plausibility. To provide a comprehensive evaluation that aligns with human perception, we conducted a rigorous user study to compare our method against others.

\begin{figure}
    \centering
    \includegraphics[width=\linewidth]{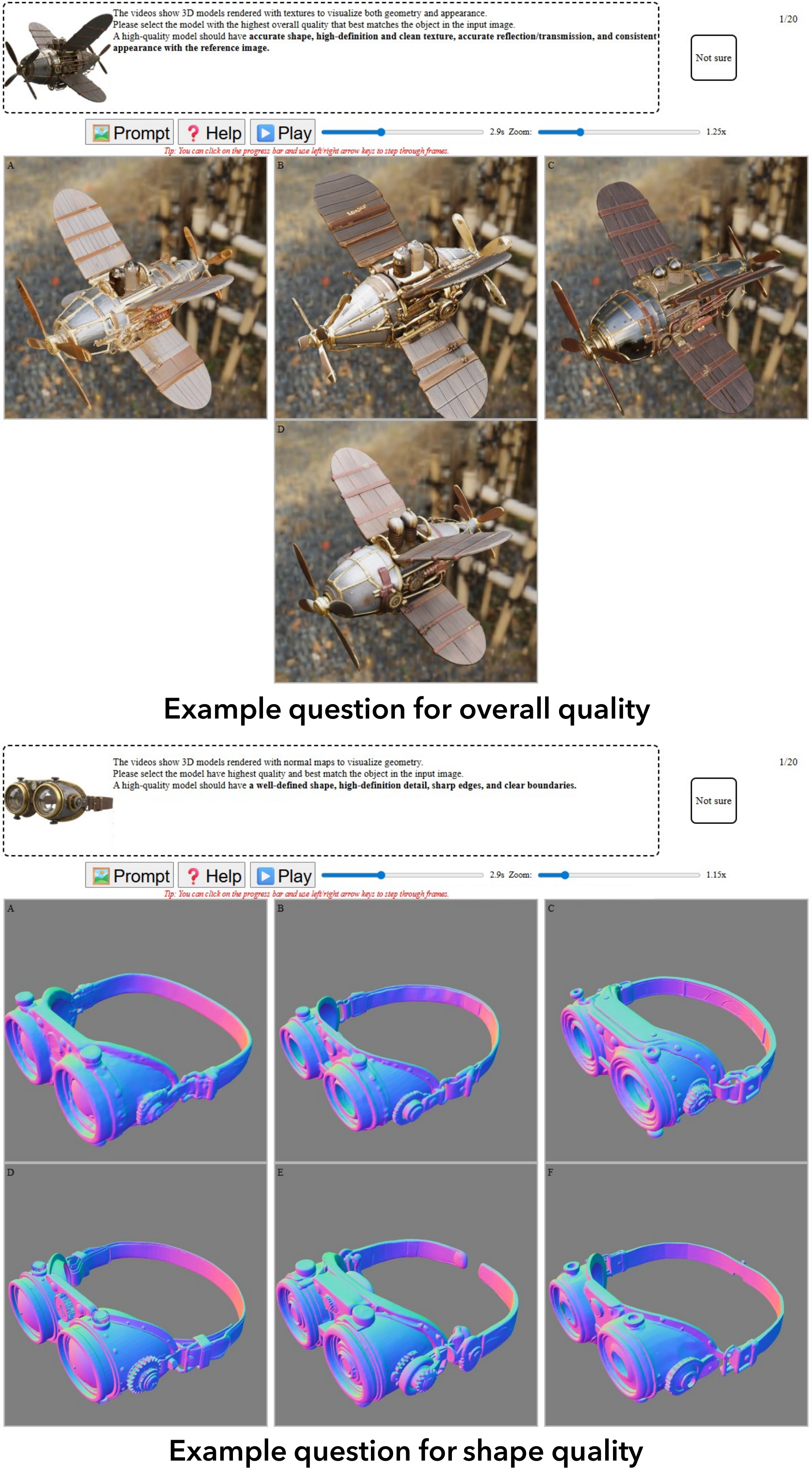}
    \vspace{-20pt}
    \caption{The interface for our user study. Participants were presented with two types of questions: one for evaluating the \emph{overall quality} of fully rendered assets (top) and another for assessing the \emph{shape quality} using normal map visualizations (bottom). The interface provided interactive controls for a thorough inspection.}
    \label{fig:user_study_ui}
\end{figure}

\paravspace
\paragraph{Study Design and Interface.}
Our study was designed to assess two critical aspects of 3D asset quality: \emph{overall quality} (combining geometry and appearance) and \emph{shape quality} (isolating geometric fidelity). Participants were presented with a series of choice questions through a custom web interface, as shown in Figure~\ref{fig:user_study_ui}.

For each question, participants were shown a reference image and a set of turntable video renderings of the 3D models generated by different methods. The interface provided interactive controls, allowing users to play, pause, scrub through the animation timeline, and zoom in to inspect details closely. This ensured that participants could perform a thorough comparison. The positions of the generated models were randomized for each question to prevent positional bias.

The study consisted of two distinct types of questions:
\begin{itemize}
    \item \textbf{Overall Quality Evaluation:} In this task, participants were shown fully textured and rendered 3D models. They were instructed to select the model with the ``highest overall quality that best matches the object in the input image.'' The evaluation criteria emphasized a holistic assessment, including accurate shape, high-definition textures, realistic material properties (reflection and transmission), and consistent appearance with the reference.

    \item \textbf{Shape Quality Evaluation:} To specifically evaluate geometric accuracy without the confounding influence of materials, this task presented the models rendered with only a normal map. Participants were asked to select the model with the best shape, focusing on criteria such as ``a well-defined shape, high-definition detail, sharp edges, and clear boundaries.''
\end{itemize}

\begin{table}[t]
    \vspace{8pt}
	\centering  
	\scriptsize
    \caption{Detailed statistics of the user study.}
    \vspace{-8pt}
	\setlength{\tabcolsep}{4pt}
	\begin{tabular}{c|cc|cc}  
		\toprule 
		\multirow{2}{*}{\textbf{Method}} & \multicolumn{2}{c|}{\textbf{Overall}} & \multicolumn{2}{c}{\textbf{Shape}} \\
        & $\textbf{Selections}\!\uparrow$ & $\textbf{Perentage}\!\uparrow$ & $\textbf{Selections}\!\uparrow$ & $\textbf{Perentage}\!\uparrow$ \\
		\midrule
        Not Sure & 4 & 2.0\% & 3 & 1.4\% \\
		\textsc{Trellis} & 13 & 6.4\% & 6 & 2.8\% \\
		Hi3DGen & -- & -- & 14 & 6.6\% \\
        Direct3D-S2 & -- & -- & 26 & 12.2\% \\
		Step1X-3D & 24 & 11.8\% & 1 & 0.5\% \\
        Hunyuan3D 2.1 & 27 & 13.3\% & 16 & 7.5\% \\
		\textbf{Ours} & \textbf{135} & \textbf{66.5\%} & \textbf{147} & \textbf{69.0\%} \\
        \midrule
        \textbf{Total} & 202 & 100\% & 213 & 100\% \\
		\bottomrule
	\end{tabular}  
	\label{tab:user_study}  
\end{table} 

\paravspace
\paragraph{Detailed Analysis.}
We recruited about 40 participants in the evaluation.
For each question, the model selected by a participant was recorded as a ``win" over the other options presented. We aggregated these results from all participants and computed a global preference rate for each method. This percentage provides a clear ranking of perceptual quality. Detailed statistics of the user study are shown in Table~\ref{tab:user_study}.

\section{More Results}

\subsection{3D Asset Reconstruction}

\begin{figure*}[p]
    \centering
    \vspace{-16px}
    \includegraphics[width=0.92\linewidth]{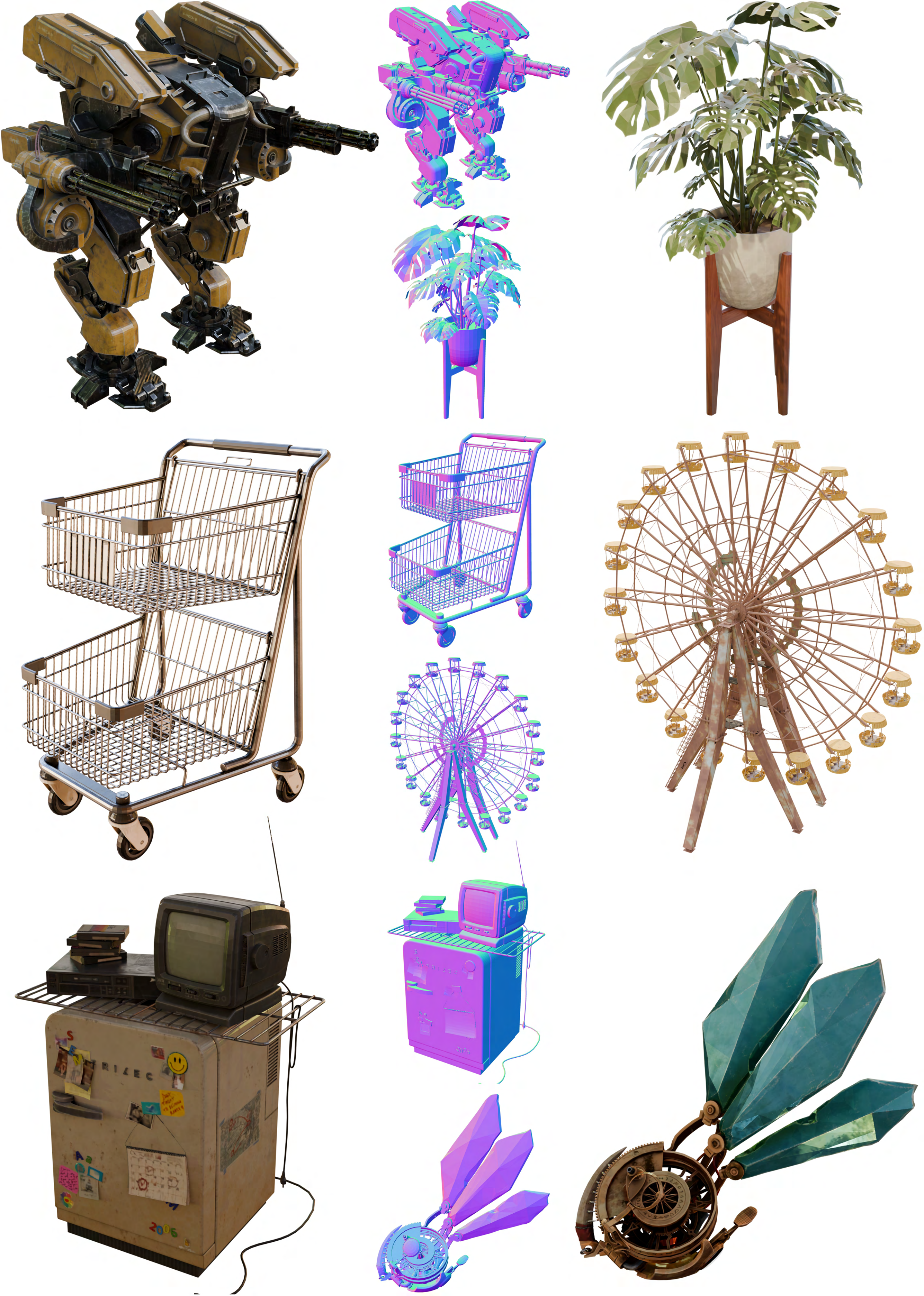}
    \vspace{-8px}
    \caption{Reconstruction results of our method. Despite highly compact, it achieves high-fidelity recovery of complex shapes and materials.}
    \vspace{-8px}
    \label{fig:results_vae}
\end{figure*}

\paragraph{Additional Reconstruction Results.}
We present additional reconstruction results of our SC-VAE in Figure~\ref{fig:results_vae}. The figure showcases the model's ability to achieve high-fidelity reconstruction across a diverse range of 3D assets. Our method successfully captures hard-surface mechanical objects (a combat mech), intricate thin structures (a shopping cart, a ferris wheel), open surfaces (a plant), words (a fridge), and complex material properties (a crystal). Despite the highly compact nature of the learned latent space, the model faithfully recovers both complex geometries, visualized via normal maps, and detailed PBR materials, shown in the final renders.

\begin{figure*}[p]
    \centering
    \vspace{-16px}
    \includegraphics[width=0.79\linewidth]{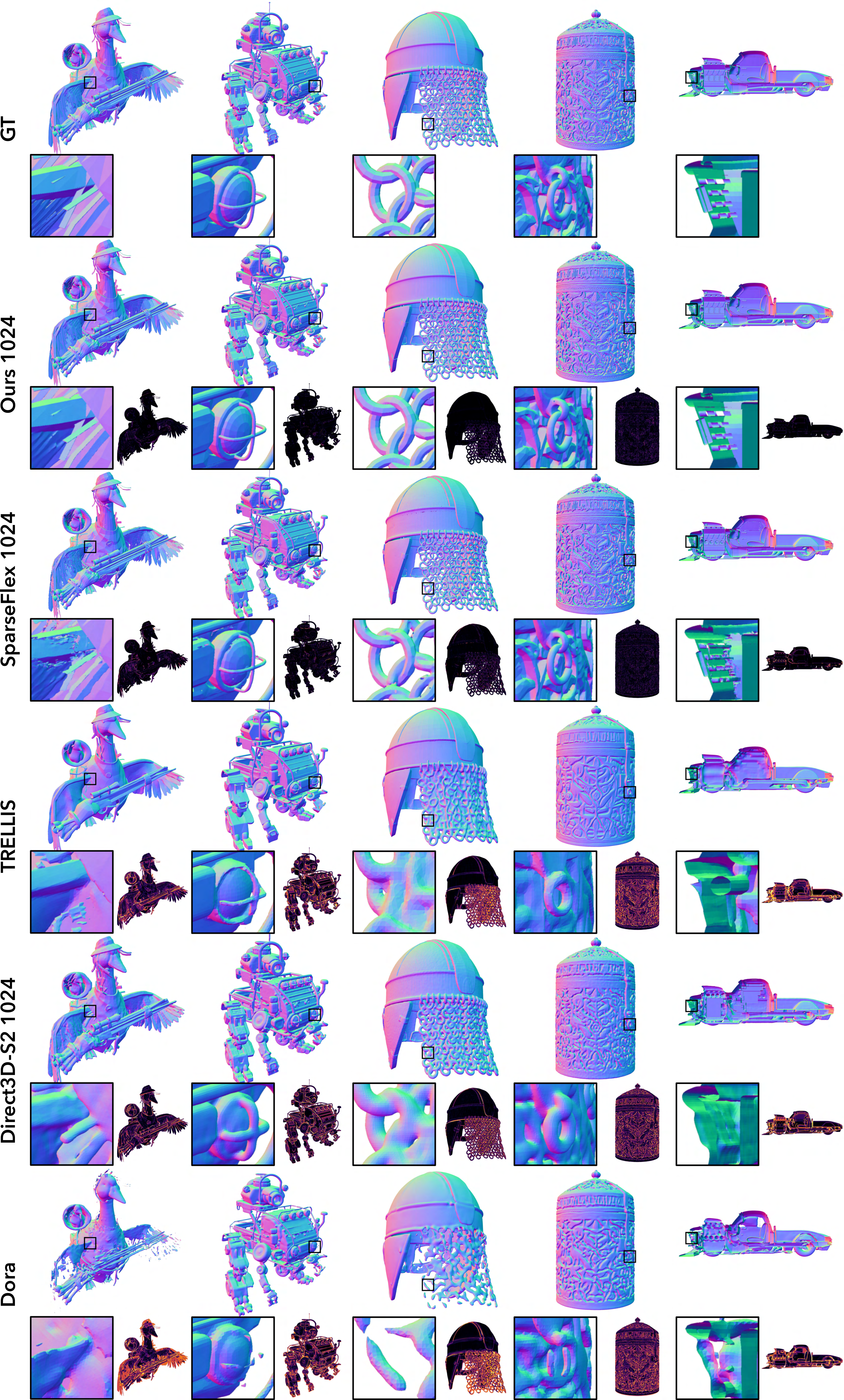}
    \vspace{-8px}
    \caption{Qualitative comparison of shape reconstruction fidelity. Error maps are shown on the bottom right.}
    \vspace{-8px}
    \label{fig:comparison_vae}
\end{figure*}

\paragraph{Additional Qualitative Comparisons.}
Figure~\ref{fig:comparison_vae} provides an extended qualitative comparison of shape reconstruction fidelity against several state-of-the-art methods. The comparison includes normal map renderings, magnified insets to highlight fine details, and corresponding error maps that visualize the deviation from the ground truth. Across all examples, our method consistently demonstrates a superior ability to preserve high-frequency geometric details. For instance, our model more accurately reconstructs the intricate chainmail links of the helmet and the sharp ornamental patterns on the decorative vessel, where other methods often produce overly smooth or blurry surfaces. Notably, as demonstrated in the final column, our method also excels at recovering enclosed internal structures, which pose a significant challenge for many surface reconstruction techniques. This high fidelity is further corroborated by the error maps, which show visibly lower reconstruction errors for our method across all examples when compared to the baselines.

\subsection{Image to 3D Asset Generation}

\begin{figure*}[p]
    \centering
    \vspace{-16px}
    \includegraphics[width=0.93\linewidth]{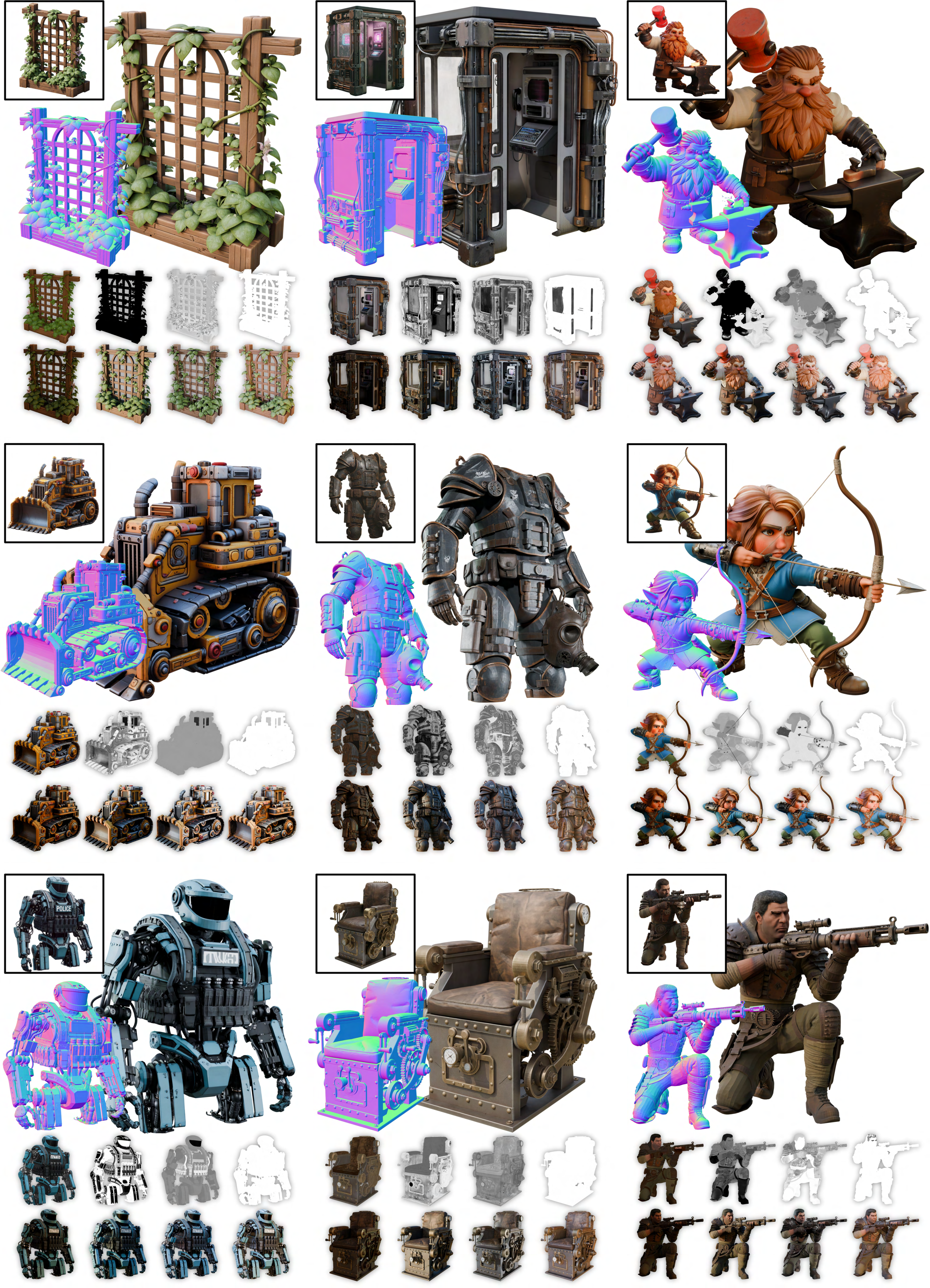}
    \vspace{-8px}
    \caption{More image-to-3D generation results of our method. PBR attrbutes and relightings are shown below. (\emph{\textbf{Best viewed with zoom}})}
    \vspace{-8px}
    \label{fig:results_more}
\end{figure*}

\paragraph{Additional Generation Results.}
We present additional qualitative results from our image-to-3D generation method in Figure~\ref{fig:results_more}. The figure demonstrates the model's versatility and robustness across a wide range of categories, including organic structures (a garden trellis with ivy), complex hard-surface machinery (a sci-fi pod, a bulldozer), and detailed characters (a dwarf blacksmith, a soldier). For each generated asset, we display the final physically-based render, the corresponding normal map to illustrate geometric detail, and a breakdown of the constituent PBR attribute maps along with relighting results. This comprehensive visualization highlights our method's ability to jointly generate not only high-fidelity geometry but also plausible PBR materials that respond correctly to novel lighting conditions.

\begin{figure*}[p]
    \centering
    \vspace{-16px}
    \includegraphics[width=0.88\linewidth]{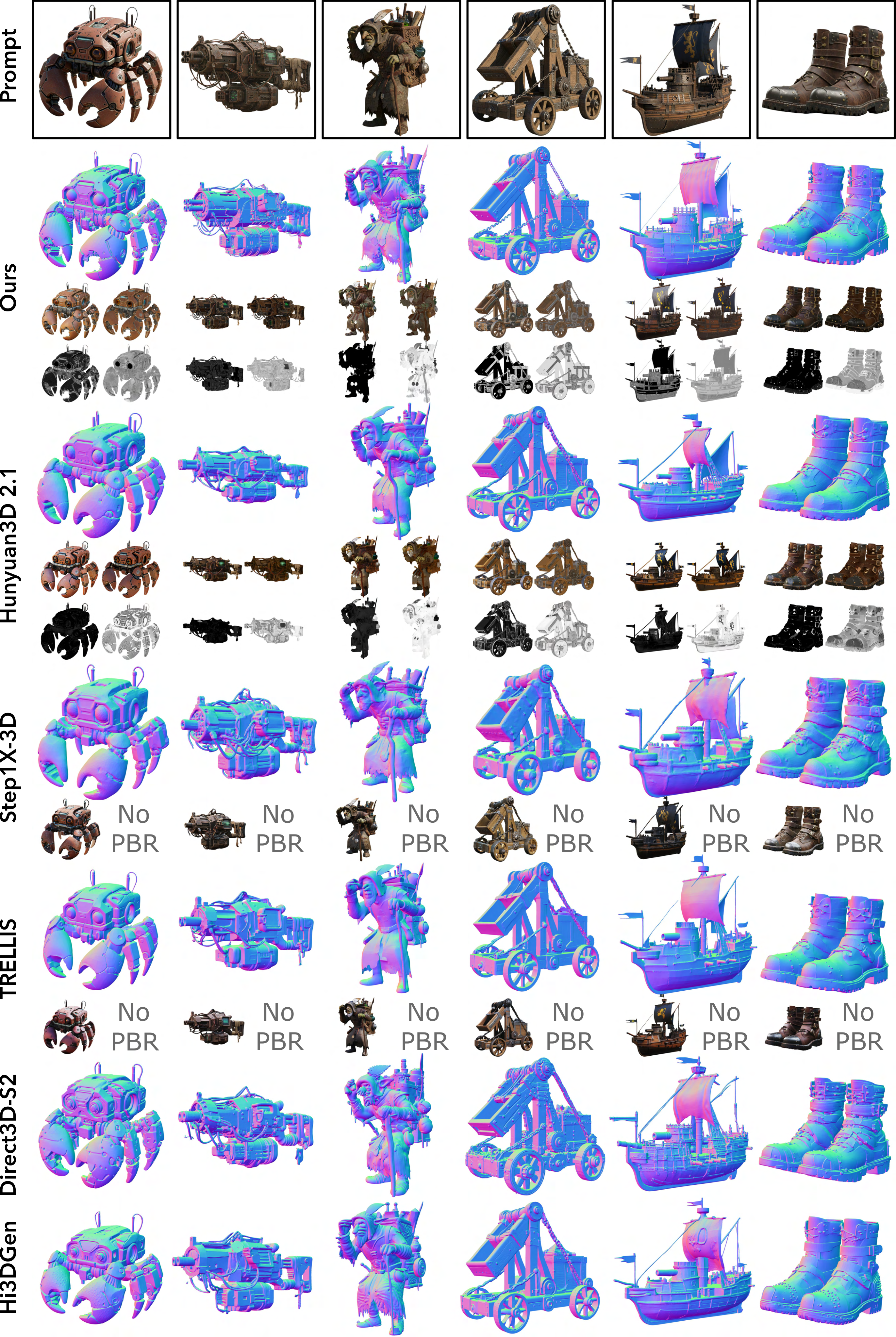}
    \vspace{-8px}
    \caption{More comparisons of image-to-3D generation results. Rendering results and PBR attrbutes (if applicable) are shown below. (\emph{\textbf{Best viewed with zoom}})}
    \vspace{-8px}
    \label{fig:comparison_more}
\end{figure*}

\paravspace
\paragraph{Additional Qualitative Comparisons.}
In Figure~\ref{fig:comparison_more}, we provide further qualitative comparisons for the image-to-3D generation task against several recent state-of-the-art methods. A primary advantage of our method is its ability to generate high-quality PBR materials, a capability not present in several baselines such as Step1X-3D, TRELLIS, Direct3D-S2, and Hi3DGen. When comparing geometric fidelity via the normal maps, our results consistently exhibit sharper and more coherent details. For example, our method more accurately captures the fine mechanical joints of the crab and the face of the character, whereas competing methods often produce results that are overly smoothed or contain noticeable artifacts. Furthermore, for methods that do produce PBR materials (Hunyuan3D 2.1), our approach generates textures that are visually more plausible and better aligned with the input prompts.

\section{Limitation Discussion and Future Work}

Despite the promising results, our method has several limitations that open avenues for future research.

First, similar to other voxel-based methods, O-Voxel's representation power is bounded by its spatial resolution. For detailed geometric features smaller than the voxel size, the Flexible Dual Grid formulation could produce aliasing artifacts.
For example, when two parallel surfaces that are very close to each other intersect the same voxel, the QEF solver, by design, will place the dual vertex at a position that minimizes the error to both surfaces, often resulting in a vertex located between them rather than accurately on one.
Similarly, the volumetric material attributes in such a voxel will be an average of the properties of both surfaces, leading to blurred appearance.

Second, we observe that the reconstructed and generated results sometimes contain small holes, though they can mostly be rectified with standard mesh post-processing techniques (e.g., hole filling). We attribute this issue to challenges in the sparse nature of our decoder, where ensuring a perfectly closed, manifold surface from the high-resolution sparse structure predicted by our decoder can be difficult. Improving the inherent stability of decoding process is an important area for improvement.

Finally, our O-Voxel is currently focused on geometry and material and it does not explicitly encode higher-level structural or semantic information. 
A significant direction for future research is to extend our representation to incorporate part-level segmentation and a graph-based topological structure. Such a structured representation would unlock an even wider range of downstream applications.

\end{document}